\documentclass[conference]{IEEEtran}

\usepackage{amsmath,amssymb}
\usepackage{newtxtext,newtxmath}
\usepackage{booktabs}
\usepackage{graphicx}
\usepackage{algorithm}
\usepackage{algpseudocode}
\usepackage{microtype}
\usepackage{placeins}
\usepackage{etoolbox}
\usepackage[hidelinks]{hyperref}

\renewcommand{\IEEEkeywordsname}{Keywords}

\patchcmd{\abstract}{\textit{\abstractname}---\relax}{}{}{
  \PackageError{mpge}{Could not remove the abstract label}{Check the IEEEtran abstract definition.}}

\providecommand{\artifactroot}{.}

\newcommand{\framework}{MPGE}
\newcommand{\method}{MPGX-CF}

\title{MPGE: A Multi-Perspective Graph Explainer for Molecular Classification Explanation}

\author{\IEEEauthorblockN{Mahtab Sarvmaili}
\IEEEauthorblockA{Dalhousie University\\
Halifax, Nova Scotia, Canada}}

\begin{document}
\maketitle
\raggedbottom

\begin{abstract}
Graph neural networks (GNNs) predict molecular properties from chemical
graph data, but predictive accuracy does not explain how graph information
supports an individual decision. A compact prediction-preserving rationale
does not necessarily reveal which changes reverse the decision or which
modifications the model tolerates. We propose the Multi-Perspective Graph
Explainer (\framework), unifying factual support, counterfactual sensitivity,
and exemplar tolerance for a frozen classifier. The factual view, originally
termed prototype (PT), seeks a compact retained edge set with the same label
and required confidence. Counterfactual (CF) explanations seek bounded
prediction-changing deletions; exemplar (EXE) explanations seek non-trivial
bounded deletions that preserve the label and confidence. A shared constrained
formulation connects prediction behavior, compactness, and edit cost, while
separate objectives generate the three views. Our graph-classification
extension of CF-GNNExplainer learns symmetric edge rankings and verifies
discrete candidates, recording unsuccessful searches. A separate BBBP
fragment backend returns RDKit-sanitized molecules. We evaluate the primary
GCN implementation on MUTAG, Mutagenicity, AIDS, COX2\_MD, and BBBP using
semantic coverage, conditional quality, stability, and runtime.
Successful factual masks retained 8.6\%--15.5\% of input edges on average across datasets; bounded counterfactual coverage was 4.8--67.6\% and exemplar preservation coverage was 98.9--100.0\%.

Exploratory controls reveal the influence of hard projection and retained
node information. Quantitative comparisons and molecular visualizations
characterize model support, sensitivity, and tolerance without treating them
as validated chemical mechanisms.
\end{abstract}


\section{Introduction}

Molecular property prediction connects chemical structure to outcomes such as
mutagenicity, biological activity, and blood--brain barrier penetration.
Graph neural networks (GNNs) provide a natural representation-learning approach
to these tasks by processing atom features and their structural relationships
\cite{wu2018moleculenet}. Predictive performance alone, however, does not reveal
which information supports an individual model decision. Explanations can
help inspect that decision and formulate structural hypotheses, although
model-faithfulness evidence must be distinguished from validated chemical
knowledge \cite{rao2022molecularxai}.

Understanding a molecular prediction requires more than identifying one
compact rationale. Factual explainers seek a subset of graph information that
supports the current prediction \cite{ying2019gnnexplainer,luo2020pgexplainer}.
Such a subset can be sufficient without identifying the changes needed to
reverse the decision. Counterfactual explainers instead search for
prediction-changing perturbations \cite{lucic2022cfgnn,bajaj2021rcexplainer}.
A successful counterfactual identifies sensitivity to a particular edit, but
does not characterize which other changes the model tolerates. These
distinctions motivate three complementary questions about the same query:
\emph{what supports the prediction, what changes it, and what can change while
the prediction is preserved?}

We address these questions through factual, counterfactual (CF), and exemplar
(EXE) explanations. The factual view seeks a compact retained edge set that
preserves the query label and required confidence. This view was originally
termed prototype (PT) in our study; we use \emph{factual} in the formal
definition and experiments because it explains an individual query rather
than a class-level prototype. The counterfactual view seeks a bounded deletion
that changes the predicted property class. The exemplar view seeks a
non-trivially modified query that retains its label and a prescribed fraction
of its confidence. Here, an exemplar is a generated query variant, not a
representative training example. A positive edit requirement excludes the
unchanged query, while the budget restricts the permitted deletions. Its
same-outcome-under-change condition is related to the neighborhood and
prediction relations of Explanation Sets \cite{fernandez2022explanation}.
Together, the views describe prediction support, sensitivity, and tolerance
within the chosen intervention space; they do not certify a chemical mechanism
or robustness to arbitrary modifications.

We propose the Multi-Perspective Graph Explainer (\framework), which unifies
these three views through a constrained formulation and a common explanation
interface. Their objectives connect prediction preservation or change with
compactness and edit cost. Prior work already couples factual and
counterfactual reasoning \cite{tan2022cf2,bajaj2021rcexplainer}; our focus is
to include the prediction-preserving edited query and evaluate all three
views under explicit output constraints. The structural implementation,
\method{}, extends the CF-GNNExplainer architecture to graph classification
\cite{lucic2022cfgnn}. It shares a frozen predictor across perspectives but
optimizes a separate symmetric edge mask for each. A ranked hard-mask search
checks the actual returned graph and records a failure when no searched
candidate meets the requested constraints. A separate BBBP fragment-search
backend returns RDKit-sanitized molecules, distinguishing chemical
representability from the structural feasibility of an edge mask.

The contributions are:
\begin{itemize}
    \item a three-view formulation connecting compact factual support,
    bounded counterfactual change, and non-trivial exemplar preservation;
    \item a graph-classification extension of CF-GNNExplainer with
    view-specific objectives, discrete feasibility checks, and explicit
    search-failure reporting;
    \item a chemistry-aware fragment-search instantiation that evaluates the
    three views in a separately sanitized molecular candidate space; and
    \item a multi-dataset evaluation that separates semantic coverage,
    conditional quality, and mask stability, with exploratory controls for
    discretization, retained node information, and edit budgets.
\end{itemize}

We evaluate the primary GCN implementation on four TU molecular graph
benchmarks---MUTAG, Mutagenicity, AIDS, and COX2\_MD---and the scaffold-held-out
BBBP dataset. Quantitative comparisons assess the three views using compatible
baselines, while visualizations show how their retained and removed structures
answer the different explanation questions. The following sections position
the framework in the literature, define its objectives and search procedures,
and present the evaluation together with the limits of molecular
interpretation.

\section{Related Work}

\subsection{Factual graph explanations}

GNNExplainer learns compact structural and feature masks that preserve
information about a model prediction \cite{ying2019gnnexplainer}.
PGExplainer amortizes mask generation over multiple inputs and also includes
size and entropy regularization \cite{luo2020pgexplainer}. These approaches
motivate our differentiable edge ranking, but our evaluation treats the final
hard subgraph, rather than its continuous surrogate, as the explanatory
object.

\subsection{Counterfactual and coupled reasoning}

CF-GNNExplainer freezes a trained GNN and learns a symmetric adjacency
sparsification mask, retaining the best discrete deletion candidate that
changes a node prediction \cite{lucic2022cfgnn}. We adapt this separation
between predictor and perturbation parameters to graph classification.
RCExplainer learns common decision logic and seeks complementary conditions
on a selected subgraph and its remainder \cite{bajaj2021rcexplainer}.
CF$^2$ directly couples factual sufficiency and counterfactual necessity
through a sparse relaxed mask \cite{tan2022cf2}. These approaches establish
coupled reasoning as an important precedent; our formulation additionally
specifies a bounded, non-trivially edited query that preserves the prediction.
Recent work formalizes post-generation graph-counterfactual minimization,
proves its computational hardness, and proposes a bounded local search
\cite{garcia2026minimization}. Our ranked-prefix scan is more restricted; it
audits feasibility along one learned edge order and does not supply a global
minimality guarantee.

CLEAR learns a generative graph-level counterfactual model, while D4Explainer
uses discrete denoising diffusion for distribution-aware counterfactual and
model-level explanations \cite{ma2022clear,chen2023d4explainer}. Their
distribution objectives address a different question from whether a discrete
graph changes a prediction. MPGE's hard checks establish the latter; its
separate fragment backend adds sanitation, not a learned molecular-manifold
constraint.

\subsection{Molecular and distribution-aware explanations}

Molecular counterfactual methods make this domain constraint concrete. MEG
trains a reinforcement-learning generator over valid molecular actions,
whereas MMACE searches model-agnostically through SELFIES mutations
\cite{numeroso2021meg,wellawatte2022molecular}. More recent counterfactual
masking replaces removed substructures with generatively completed fragments
to reduce out-of-distribution evaluation artifacts
\cite{janisiow2026counterfactual}. ICExplainer likewise uses a learned graph
distribution, and ATEX-CF (ICLR 2026) expands node-level counterfactual search
to both edge additions and deletions
\cite{chen2026icexplainer,zhang2026atexcf}. Our deletion-only
fragment search explores a restricted candidate family. It instantiates
all three views under one frozen predictor without claiming the broader
generation capabilities of these methods.

Recent work also changes the explanatory object or its optimization space.
MAGE generates model-level graph explanations using motifs, including
molecular examples \cite{yu2025mage}. IDEA aligns instance-level explanations
in a prototypical representation space rather than only matching graph
labels \cite{yin2026idea}. Neither a class-level generated motif nor a latent
prototype is equivalent to the compact query-specific factual mask termed PT
in the original version of our study. This distinction matters because
prediction preservation alone does not identify a unique explanatory
structure.

\subsection{Exemplars and evaluation}

The same-outcome-under-change perspective motivates the invariance constraint
used by EXE. Explanation Sets formalize several example-based views within a common neighborhood and
prediction relation \cite{fernandez2022explanation}. We adapt that structure
to a generated subgraph and require a positive edit cost, preventing the
original graph from being returned as its own exemplar. This definition is
stated explicitly because other explainability work may use \emph{exemplar}
for a representative observed case rather than a transformed query.

GraphFramEx separates explanation user needs and combines necessity and
sufficiency evidence \cite{amara2022graphframex}. Other systematic analyses
show that faithfulness does not imply stability
\cite{agarwal2022probing}, and explanation masks can be fragile to small
perturbations \cite{li2024fragile}. We consequently report semantic success, coverage,
size or edit cost, stability, and runtime separately instead of collapsing
them into one unqualified accuracy score.
Recent work additionally constructs degenerate GNN explanations that standard
faithfulness metrics can miss \cite{azzolin2026degenerate}. Although that
analysis targets self-explainable GNNs rather than our post-hoc setting, it
motivates interpreting favorable fidelity together with structure and stability
diagnostics rather than as a complete explanation-quality certificate.
Accordingly, MPGE is positioned as a query-level, three-view formulation with
explicit hard-output evaluation, not as the first combination of explanation
perspectives or a distribution-certified molecular generator.

\section{Problem Formulation}

Let $G=(V,E,X)$ be an attributed undirected graph and let a trained, frozen
binary classifier $f$ return a logit. Define
\begin{equation}
    p(G)=\sigma(f(G)),\qquad
    y(G)=\mathbb{I}[p(G)\geq 0.5].
\end{equation}
For the query graph $G$, fix $y=y(G)$ and define, for any candidate $H$,
$p_y(H)=p(H)$ if $y=1$ and $p_y(H)=1-p(H)$ otherwise. Thus $p_y$ always
refers to the original query class, even after a flip. An edit set $\Delta$
induces $T(G,\Delta)$. Structural candidates have $H=(V,E_H,X)$ with
$E_H\subseteq E$; molecular candidates are generated connected fragments.
Structural validity requires symmetric deletion of existing edges with
unchanged nodes and features. Molecular validity additionally requires the
fragment-generation and RDKit sanitation checks. Neither predicate establishes
distributional plausibility or a physical property. The evaluated structural
queries have $M=|E|\geq1$.

Let $A_F(H)$ be a backend-specific factual size, $C(\Delta)$ the edit cost,
and $\mathcal D_B$ the permitted bounded edits. Structurally,
$A_F(H)=|E_H|$, $C(\Delta)=|E\setminus E_H|$, and
$\mathcal D_B=\{\Delta:1\leq C(\Delta)\leq B\}$. The fragment backend uses
retained heavy atoms for $A_F$, deletion steps for $C$, and an additional
removed-heavy-atom fraction limit in $\mathcal D_B$.

\subsection{Factual explanation}

A factual explanation is a compact retained graph $S_F\subseteq G$.
For the structural backend, $S_F=(V,E_F,X)$ retains every query node and
feature and selects only $E_F\subseteq E$; its compactness is therefore
\emph{edge} compactness. The chemistry-aware backend instead returns a
sanitized connected fragment under the same prediction constraint but a
different size cost:
\begin{equation}
\begin{aligned}
    \min_{S_F}\quad & A_F(S_F) \\
    \text{s.t.}\quad
    & y(S_F)=y,\\
    & p_y(S_F)\geq \tau_F,\\
    & S_F\in\mathcal S_F(G),\quad \operatorname{Valid}(S_F)=1.
\end{aligned}
\label{eq:factual}
\end{equation}
Here $\mathcal S_F(G)$ contains non-empty retained-edge graphs for the
structural backend and generated connected fragments with at least two atoms
for the molecular backend. The former excludes a zero-edge output but retains
all node information. It also contains the full query, so factual feasibility
is expected for a valid non-empty structural query; compactness, rather than
coverage alone, distinguishes factual outputs. The confidence thresholds are
relative to the unmasked prediction,
$\tau_F=\tau_E=\rho p_y(G)$, with $\rho$ fixed before test evaluation.

\subsection{Counterfactual explanation}

A counterfactual is a minimum-cost valid intervention that changes the
decision:
\begin{equation}
\begin{aligned}
    \min_{\Delta_C}\quad & C(\Delta_C) \\
    \text{s.t.}\quad
    & y(T(G,\Delta_C))\neq y,\quad \Delta_C\in\mathcal D_B,\\
    & \operatorname{Valid}(T(G,\Delta_C))=1.
\end{aligned}
\label{eq:counterfactual}
\end{equation}
If $m$ is a keep mask, the deletion cost is proportional to
$\lVert 1-m\rVert_1$, not $\lVert m\rVert_1$. This direction is essential:
penalizing the keep mask would reward deleting more edges.

\subsection{Exemplar explanation}

An exemplar is a non-trivial, bounded change under which the decision and a
minimum confidence are preserved:
\begin{equation}
\begin{aligned}
    \max_{\Delta_E}\quad & U(G,T(G,\Delta_E)) \\
    \text{s.t.}\quad
    & y(T(G,\Delta_E))=y,\\
    & p_y(T(G,\Delta_E))\geq \tau_E,\\
    & \Delta_E\in\mathcal D_B,\\
    & \operatorname{Valid}(T(G,\Delta_E))=1.
\end{aligned}
\label{eq:exemplar}
\end{equation}
Here $U$ measures a specified structural difference from the query. Positive
cost excludes a no-op and the budget restricts the edit family; neither
guarantees meaningfulness to a domain expert. We use removed-edge fraction
for masks and one minus Morgan-fingerprint similarity for molecular fragments.
Factual and exemplar views share a prediction-preservation condition but
optimize different quantities under different candidate constraints. The
implementation searches restricted families and does not certify global optima.

\section{MPGE Framework and MPGX-CF Architecture}

\subsection{Frozen predictor and perturbation generator}

Let $f_W$ be the trained graph classifier and let $g_P$ be its perturbation
counterpart. As in CF-GNNExplainer \cite{lucic2022cfgnn}, $g_P$ uses the frozen
weights $W$ of $f_W$ and learns only structural gates $P$. The predictor is in
evaluation mode: dropout is disabled and BatchNorm running statistics are
fixed. Our implementation
assigns one logit $z_e$ to each of the $M$ unique undirected edges and mirrors
$m_e=\sigma(z_e)$ onto both directed message-passing entries. Consequently,
$\widetilde A=A\odot m$ is symmetric, cannot introduce a non-edge, and does not
gate the self-messages added by the GCN. The GCN normalization is recomputed on
$\widetilde A+I$ during every forward pass. This is the graph-classification
analogue of the CF-GNNExplainer generating model, rather than a second
predictor trained from data.

The gate logits are initialized as $z_e=1+\epsilon_e$, with
$\epsilon_e\sim\mathcal{N}(0,0.01^2)$, giving initial soft weights near
$\sigma(1)$ rather than unit weights. Let $q_y(m)$ denote the probability
assigned by $g_P$ to the query class
under the soft mask, $\overline m=M^{-1}\sum_e m_e$, and $H(m)$ denote mean
binary entropy. We optimize a separate $P$ for each requested view; sharing the
predictor does not imply sharing an explanation mask.

\subsection{Perspective-specific objectives}

The factual relaxation is
\begin{equation}
    L_F =
    \lambda_p[\tau_F-q_y(m)]_+
    +\beta_1\overline m+\beta_2\overline{m^2},
\end{equation}
where $\overline m$ is the mean retained mass. The counterfactual relaxation
reverses the decision condition and penalizes removed rather than retained
mass:
\begin{equation}
    L_C =
    \lambda_p[q_y(m)-0.49]_+
    +\beta_1\overline{(1-m)}+\beta_2\overline{(1-m)}^2.
\end{equation}
The exemplar relaxation preserves confidence while encouraging a
non-trivial deletion up to budget $b$:
\begin{equation}
\begin{split}
    L_E={}&\lambda_p[\tau_E-q_y(m)]_+
    -\beta_1\overline{(1-m)}\\
    &+\lambda_b[\overline{(1-m)}-b]_+^2
    +\beta_2H(m).
\end{split}
\end{equation}
Here $[a]_+=\max(a,0)$, $b$ is the soft edit-fraction limit, and
$\lambda_p=\lambda_b=8$ in the locked configuration. The counterfactual
surrogate targets original-class probability below 0.49; actual hard success
uses the classifier's 0.5 decision rule. The factual quadratic term averages
squared gates, whereas the counterfactual term squares the mean edited mass;
the exemplar term is entropy regularization, not an elastic-net penalty.
These losses are differentiable surrogates, not the final explanations. After
optimization, the algorithm orders the unique undirected gates and evaluates
hard masks in ranked order. It selects the smallest
sufficient factual mask, the first prediction-changing deletion set, or the
largest in-budget confidence-preserving exemplar deletion set. If no hard
candidate satisfies its definition, the method returns a failure. This avoids
the soft--hard mismatch that can arise when the optimized mask is thresholded.
This deterministic feasibility audit extends CF-GNNExplainer's best-candidate
tracking to all three perspectives. These selections are extremal only within
the ranked candidate sequence; neither the edge-mask scan nor the bounded
fragment beam proves a global optimum of Eqs.~(\ref{eq:factual})--(\ref{eq:exemplar}).

\begin{algorithm}[t]
\caption{Multi-Perspective CF-GNN Explanation for Graph Classification}
\label{alg:multi-perspective}
\footnotesize
\begin{algorithmic}[1]
\Require graph $G=(V,E,X)$ with $M=|E|\geq1$; logit predictor $f_W$;
view $v\in\{F,C,E\}$; $\rho$; fraction $b$; absolute cap $B_{\max}$;
$\beta_1,\beta_2,\lambda_p,\lambda_b$; steps $K$; learning rate $\eta$; seed $s$
\Ensure hard explanation $G_v$ or an explicit failure
\State set predictor to evaluation mode; freeze every parameter in $W$
\State $y\gets y(G)$; $\tau\gets\rho p_y(G)$
\State $B\gets\min\{M,B_{\max},\max(1,\lfloor bM\rfloor)\}$
\State create one logit $z_e$ for every unique undirected edge $e\in E$
\State initialize $z$ from seed $s$
\For{$t=1,\ldots,K$}
    \State $m\gets\sigma(z)$; mirror $m$ onto both edge directions
    \State $q_y\gets p_y(G\odot m)$ using $\sigma(f_W(\cdot))$
    \State update $z$ by Adam to \emph{minimize} $L_v(q_y,m)$
\EndFor
\State $m\gets\sigma(z)$ after the final update
\If{$v=F$}
    \State order edges by decreasing $m_e$
    \State $\mathcal C\gets$ retained-edge prefixes from smallest to largest
    \State $G_v\gets$ first $H\in\mathcal C$ with
    $y(H)=y$ and $p_y(H)\geq\tau$
\ElsIf{$v=C$}
    \State order edges by increasing $m_e$
    \State $\mathcal C\gets$ deletion prefixes of size $1,\ldots,B$
    \State $G_v\gets$ first $H\in\mathcal C$ with $y(H)\neq y$
\Else
    \State order edges by increasing $m_e$
    \State $\mathcal C\gets$ deletion prefixes of size $1,\ldots,B$
    \State $G_v\gets$ $H\in\mathcal C$ with the largest deletion prefix and
    $y(H)=y$ and $p_y(H)\geq\tau$
\EndIf
\If{no candidate satisfies the hard constraints}
    \State \Return \textsc{Failure} with the violated constraint
\EndIf
\State \Return $G_v$, its hard mask, probabilities, edits, and provenance
\end{algorithmic}
\end{algorithm}

Candidates are symmetric deletions of existing undirected edges, satisfying
structural validity by construction. Algorithm~\ref{alg:multi-perspective}
specifies the proposed structural graph-classification extension of
CF-GNNExplainer, not the fragment backend or the baselines. It
optimizes only unique-edge gates and uses gradient descent through Adam; it
does not optimize dense non-edge entries or update the predictor. Most
importantly, the returned object is selected by an explicit hard feasibility
scan rather than by thresholding a soft mask and assuming that its semantics
survive discretization. A failure means that no searched candidate was feasible,
not that no explanation exists in the full edit space.

The structural backend masks edges but retains the full node set. The GCN also
adds a self-loop at every node. Consequently, a one-edge factual mask can still
use every node's features through self-messages and should be interpreted as
an edge-sufficiency result, not as a two-node induced subgraph or a chemically
complete fragment. We therefore report retained edges together with isolated
nodes and connected components, and reserve molecular-fragment language for
the chemistry-aware backend below.

\subsection{Chemistry-aware fragment search}

Arbitrary deletion masks can leave atom sets that do not describe a useful
molecule. For molecular graphs with SMILES provenance, we therefore implement
a second search space. At each step the method cuts a non-ring single bond,
retains a connected component, restores implicit hydrogens through
canonicalization, and accepts the candidate only if RDKit sanitization
succeeds. Duplicate canonical SMILES are removed.

The search advances for at most three deletion steps. At each depth, at most
100 distinct candidates are retained, ordered by decreasing heavy-atom count
and then canonical SMILES. Every retained fragment has at least two atoms;
the unchanged query is not generated. Factual selection minimizes retained
heavy atoms, breaking ties by fewer steps and greater query similarity.
Counterfactual selection minimizes deletion steps, then removed heavy atoms,
then prefers greater similarity. Exemplar selection maximizes fingerprint
distance, breaking ties by more steps and fewer retained heavy atoms. Only
counterfactual and exemplar candidates are restricted to removal of at most
35\% of the query's heavy atoms. Morgan fingerprints use radius two and the
recorded RDKit default bit-vector settings.

Sanitization checks representability and valence consistency
\cite{rdkitbook}; it does not establish synthetic accessibility, likely
occurrence, or a physically changed property. This backend adds chemical
checks but is not a learned in-distribution generator.

\subsection{Predictive model}

The primary classifier is a three-layer GCN \cite{kipf2017gcn}. With
$\bar A=A+I$ and
$\bar D_{ii}=\sum_j\bar A_{ij}$, a layer is
\begin{equation}
Z^{(\ell+1)}=
\bar D^{-1/2}\bar A\bar D^{-1/2}H^{(\ell)}W^{(\ell)}
+a^{(\ell)},
\end{equation}
where $a^{(\ell)}$ is a bias. Each convolution is followed by BatchNorm,
ReLU, and training-only dropout. Node representations are sum-pooled and
passed to a linear binary-logit head; sigmoid obtains probabilities.
Categorical node embeddings are summed column-wise; continuous node features
use a linear encoder. The primary GCN consumes node features and topology,
not edge labels or distance attributes. The GINE configuration used in
validation sensitivity checks additionally encodes edge features; it is not
a completed held-out architectural robustness study.

\section{Experimental Protocol}

\subsection{Research questions and evaluation scope}

We organize the evaluation around five questions. RQ1 measures whether each
hard output satisfies its factual, counterfactual, or exemplar definition.
RQ2 evaluates compactness, faithfulness, and edit cost conditional on
success while retaining failures in the coverage denominator. RQ3 tests
whether the views select different structures on jointly successful cases.
RQ4 measures repeatability across explainer seeds and sensitivity to the
representation and edit budget. RQ5 tests how much counterfactual coverage
changes when rankings
learned by different objectives receive the same hard-prefix projection.
This replay does not isolate every difference in optimization or training budget.

\subsection{Datasets, features, and splits}

Table~\ref{tab:datasets} summarizes the five evaluated datasets.

MUTAG and Mutagenicity concern mutagenicity classification
\cite{debnath1991structure,kazius2005derivation}; AIDS provides a molecular
activity classification benchmark through the IAM repository
\cite{riesen2008iam}. COX2\_MD derives from COX2 activity data
\cite{sutherland2003spline} and its distance-graph construction
\cite{kriege2012subgraph}: after removing explicit hydrogens, complete graphs
carry interatomic-distance attributes and bond-type labels, including a
no-bond category. Removing a COX2\_MD graph edge is consequently not
necessarily removing a chemical bond. We load all four from the official TU
archives \cite{morris2020tudataset} and preserve source indices, categorical
node and edge labels, and continuous attributes. Continuous attributes are
standardized using training-fold statistics only. Stored edge labels and
attributes are available to the GINE validation configuration but not consumed
by the primary GCN. They are not explanation annotations; consequently, we
do not report edge-mask accuracy on these datasets. For each TU dataset, one deterministic
stratified ten-fold partition is created with master seed 20260911. In outer
fold $k$, fold $k$ is test, fold $(k+1)\bmod 10$ is validation, and the
remaining folds are training. Every graph is therefore tested exactly once
per predictor seed.

Predictor metrics use all ten outer folds. Explanation generation is complete
for all ten folds of MUTAG, Mutagenicity, and COX2\_MD and for folds 0--7 of
AIDS; the latter contains 1,600 distinct held-out graphs per predictor seed.
The reporting configuration excludes the incomplete AIDS fold 8 as a whole
and the unstarted fold 9, independent of their outcomes. BBBP uses its single
locked scaffold test set. Thus every explanation table and plot uses only
complete dataset--fold--predictor cells, with all attempted graphs and failed
explanations retained inside those cells.

BBBP is loaded from the processed MoleculeNet/PyG collection
\cite{wu2018moleculenet,martins2012bayesian}. We record
the source row, canonical SMILES, Bemis--Murcko scaffold, label, and any
additional loader rejection. Table~\ref{tab:datasets} counts accepted graphs,
not rejected or upstream-filtered source molecules. A deterministic
stratified-group split groups equal scaffold strings, including empty strings
for acyclic molecules. It selects test and validation folds closest to the
target sizes and overall class balance, yielding approximately 80\%, 10\%,
and 10\% training, validation, and test partitions with zero scaffold overlap.
Dataset manifests contain SHA-256 hashes of all available raw and processed
files, feature schemas, graph identifiers, and split membership.

\begin{table*}[t]
\centering
\caption{Datasets and locked evaluation protocol. $|E|$ counts each
undirected edge once. N:L and E:L give node/edge categorical-label
cardinalities; N:A and E:A give continuous-attribute dimensions. BBBP uses
MoleculeNet atom/bond categorical features. GT denotes a documented
edge-level explanation mask; none of these five datasets provides one.
Features describe stored data, not all inputs used by GCN. AIDS explanation
reporting uses folds 0--7; predictor reporting uses all ten folds.}
\label{tab:datasets}
\small
\begin{tabular}{lrrrrllc}
\toprule
Dataset & $N$ & Pos. & $|V|$ & $|E|$ & Features & Split & GT \\
\midrule
MUTAG & 188 & 0.66 & $17.9 \pm 4.6$ & $19.8 \pm 5.7$ & N:L7; E:L4 & Stratified 10-Fold & no \\
Mutagenicity & 4337 & 0.45 & $30.3 \pm 20.1$ & $30.8 \pm 16.8$ & N:L14; E:L3 & Stratified 10-Fold & no \\
AIDS & 2000 & 0.80 & $15.7 \pm 13.7$ & $16.2 \pm 15.0$ & N:L38; N:A4; E:L3 & Stratified 10-Fold & no \\
COX2\_MD & 303 & 0.49 & $26.3 \pm 2.5$ & $335.1 \pm 65.3$ & N:L7; E:L5; E:A1 & Stratified 10-Fold & no \\
BBBP & 2039 & 0.77 & $24.1 \pm 10.6$ & $26.0 \pm 11.7$ & Atom/Bond Categorical & Scaffold 80/10/10 & no \\
\bottomrule
\end{tabular}

\end{table*}

\subsection{Predictors and eligibility}

The frozen predictor is a three-layer GCN, with 64 hidden units per
layer, BatchNorm, ReLU, dropout 0.2, sum pooling, and a binary logit head.
Training uses Adam, class-weighted binary cross-entropy with
the positive weight estimated from the training partition, learning rate
$10^{-3}$, weight decay $10^{-4}$, batch size 64, at most 200 epochs, and early
stopping after 30 epochs without improvement in validation log loss. Model
seeds are 11, 29, and 47.

We report test AUROC, AUPRC, balanced accuracy, F1, Brier score, expected
calibration error, log loss, and the selected epoch before reporting any
explanation result. All predictions in the locked explanation population are
eligible for model-faithfulness analysis because the target is the frozen
model's prediction. The displayed dataset-level dagger is assigned when the
mean balanced accuracy is below 0.55 or mean AUROC below 0.60. It cautions
against application-utility interpretation; explanation eligibility is not
conditioned on predictive correctness.

\subsection{Hyperparameter selection and budgets}

No test graph participates in selection. We use at most 24
deterministically selected, class-balanced validation graphs per fold with
predictor seed 11 and explainer seed 101 to evaluate the complete grid
$\beta_1\in\{0.1,0.001,0.0001\}$ and
$\beta_2\in\{0.1,0.01,0.001\}$ for each perspective. For factual and
counterfactual views, $\beta_1$ and $\beta_2$
both regularize retained mass for factuals and edited mass for counterfactuals;
for exemplars they control
the edit reward and binarization penalty. One global configuration per
perspective is selected lexicographically by: (i) maximum
hard-constraint success, (ii) minimum retained fraction for factuals, minimum
edit count for counterfactuals, or maximum feasible edit fraction for
exemplars, (iii) maximum retained query-class confidence, and (iv) minimum
runtime. Dataset-level validation means are macro-averaged so Mutagenicity
does not dominate smaller domains. We additionally evaluate edit-fraction
budgets 0.05, 0.10, 0.20, 0.30, and 0.35. The default hard budget is
$B=\min\{|E|,8,\max(1,\lfloor0.35|E|\rfloor)\}$ and $\rho=0.8$.
The minimum-one clamp applies to non-empty structural graphs, relaxing the
fractional limit when a graph is too small to admit one deletion.
Explainer seeds are 101,
211, and 307. The \method{} gate optimizer runs for 200 steps with Adam at
learning rate 0.03. CF$^2$ follows the official 500-step, 0.05 learning-rate
graph configuration. PGExplainer is trained within each training partition
for 20 epochs at learning rate 0.003; this training time is excluded from
per-query explanation timings.

\subsection{Comparison methods}

All compatible methods use the same predictor checkpoints, graph identifiers,
splits, and output-feasibility criteria, but their native discretization rules
differ. GNNExplainer and fold-trained PGExplainer are
factual baselines \cite{ying2019gnnexplainer,luo2020pgexplainer}. A
graph-classification adaptation of CF-GNNExplainer is evaluated only for
counterfactual search \cite{lucic2022cfgnn}. It uses the published
negative-NLL prediction term while the thresholded label remains unchanged,
plus $\beta=0.5$ times the differentiable deletion distance. Following the
authors' released run commands, its symmetric gates are optimized for 500 SGD
steps at learning rate 0.1 with Nesterov momentum 0.9; the best hard,
in-budget flip is retained. CF$^2$ uses the authors' joint
$L_1+\lambda[\alpha L_f+(1-\alpha)L_c]$ objective, initialization, 0.5 mask
threshold, and published graph defaults ($\lambda=1000$, $\alpha=0.6$,
$\gamma=0.5$); the same learned mask is evaluated as a factual subgraph and
its complement for necessity \cite{tan2022cf2}. A one-edge-deletion greedy
ranking is the exemplar control. Matched random rankings are evaluated for all
three perspectives.

To assess this discretization confound, we replay stored counterfactual
rankings through one identical hard-prefix search on a deterministic,
class-balanced subset of 24 held-out graphs per dataset (12 per original
predicted class), with all three GCN seeds and explainer seed 101. The subset
contains only graph IDs with complete records for all compared rankings. We
also evaluate the proposed soft mask at a fixed 0.5 threshold, and test both
a node-feature-only zero-edge input and an induced-node version of each
selected factual explanation. These are exploratory controls, not a
replacement for the locked full-population experiment.

For BBBP, we additionally report a separate sanitized fragment-deletion
search. RDKit validity establishes graph representability, not
synthesizability, and these chemistry-aware results are not pooled with the
structural edge-mask comparison.

\subsection{Outcomes and statistical analysis}

Semantic success is the hard label/confidence predicate of the requested view.
Factual quality uses retained-edge fraction and comprehensiveness,
$p_y(G)-p_y(G\setminus E_F)$, where the complement retains all query nodes.
Counterfactual quality uses deleted-edge count and original-class probability
drop, $p_y(G)-p_y(H)$. Exemplar quality uses deleted-edge fraction and
retained-confidence ratio, $p_y(H)/p_y(G)$. Structural edits are deletion-only;
no edge-addition outcome or geometric boundary-distance estimate is inferred.
Components and isolated nodes are counted on the full retained node set without
predictor-added self-loops. Probability-shift diagnostics use $|p(H)-p(G)|$.
Centrality diagnostics are root-mean-square nodewise changes in
NetworkX-normalized betweenness and unnormalized harmonic centrality, not
chemical importance scores.

Stability averages pairwise Jaccard similarity across successful explainer
repetitions for the same graph--model pair, using retained edges for factuals
and deleted edges for other views. Units with fewer than two successful
repetitions supply no stability value. Cross-perspective Jaccard overlap
compares factual retained edges with counterfactual or exemplar deleted edges
on complete successful triplets. Runtime is wall time per query and view,
including failures, and is distinct from the sum across views. Timing depends
on implementation and hardware; it is not a universal complexity comparison.

Failed searches remain in the denominator and retain a machine-readable reason.
Explainer repetitions are averaged within each graph--model pair. Main-text
continuous outcomes are reported as mean $\pm$ standard deviation over these
analysis units, while success is reported as a percentage of all attempts,
including failed searches. Descriptive intervals in the supplementary metric
records use 10,000 hierarchical bootstrap resamples over fold, model seed,
and graph as applicable; new timing intervals use the same hierarchy.
Paired effect intervals instead resample unique held-out graphs after averaging
their repeated predictor-seed differences. Method tests use these paired
graph-level differences on shared examples, Wilcoxon signed-rank tests,
and Holm correction within each dataset--perspective--metric family. Datasets
are separate domains; no micro-average across datasets is used. Shared fitted
models induce dependence among held-out graphs, so intervals and paired tests
are descriptive of the evaluated population, not guarantees of generalization
to new training samples.

\subsection{Qualitative selection and reproducibility}

The overview prefers complete successful \method{} triplets selected by
minimum distance to dataset medians of mean edit count and mean retained
query-class probability. If none exist, it retains cases with the most
successful views before applying the same rule. This deterministic
illustration rule is not a claim of prospective preregistration. The
appendices provide enlarged views and a failure audit.
TU panels display documented categorical labels and are not
described as sanitized molecules. Every panel is regenerated from its graph ID
and hard mask and records fold, predictor seed, explainer seed, probabilities,
edit count, and failure status.

The reported experiment is staged as predictor training, validation-only
selection, locked test explanation, and artifact generation. Runs
are resumable by dataset, fold, architecture, predictor seed, explainer seed,
graph, method, perspective, budget, and protocol hash. The local reproducibility
package records the
resolved configuration, environment, source commit (or SHA-256 source hashes
when commit metadata are unavailable), checkpoints, histories, split
manifests, failures, and long-form CSV/Parquet output. All dependencies
are installed in project-local UV environments. The arXiv source archive
contains the manuscript, rendered tables and figures, a claim audit,
supplementary metric summaries, and compact provenance CSVs, but not the large raw
experiment directory; a public accession for the latter has not been assigned.

\section{Results and Discussion}
\label{sec:results}

\subsection{Predictor performance and calibration}

Predictive results establish the context for the explanation study. A dagger
marks a dataset-level mean that triggers the weak-predictor rule; its
explanations characterize the frozen model, not established application utility.

\begin{table*}[t]
\centering
\caption{Primary GCN out-of-fold or scaffold-held-out predictive performance.
Values are mean $\pm$ standard deviation over outer-fold/model-seed runs.}
\label{tab:model-performance}
\scriptsize
\begin{tabular}{llcccc}
\toprule
Dataset & Architecture & Balanced accuracy & AUROC & AUPRC & F1 \\
\midrule
AIDS & GCN & $0.971 \pm 0.014$ & $0.985 \pm 0.015$ & $0.991 \pm 0.012$ & $0.989 \pm 0.005$ \\
BBBP & GCN & $0.698 \pm 0.014$ & $0.797 \pm 0.010$ & $0.942 \pm 0.008$ & $0.907 \pm 0.008$ \\
COX2\_MD$^{\dagger}$ & GCN & $0.519 \pm 0.069$ & $0.561 \pm 0.133$ & $0.586 \pm 0.099$ & $0.522 \pm 0.181$ \\
MUTAG & GCN & $0.688 \pm 0.160$ & $0.836 \pm 0.102$ & $0.929 \pm 0.043$ & $0.835 \pm 0.062$ \\
Mutagenicity & GCN & $0.781 \pm 0.024$ & $0.856 \pm 0.022$ & $0.819 \pm 0.029$ & $0.757 \pm 0.027$ \\
\bottomrule
\end{tabular}

\end{table*}

\begin{table*}[t]
\centering
\caption{Predictive calibration and selected training epoch; mean $\pm$
standard deviation over the same outer-fold/model-seed runs.}
\label{tab:model-calibration}
\scriptsize
\begin{tabular}{llcccc}
\toprule
Dataset & Architecture & Brier & ECE & Log loss & Best epoch \\
\midrule
AIDS & GCN & $0.015 \pm 0.006$ & $0.023 \pm 0.007$ & $0.077 \pm 0.050$ & 37.8 \\
BBBP & GCN & $0.122 \pm 0.005$ & $0.094 \pm 0.010$ & $0.466 \pm 0.035$ & 10.3 \\
COX2\_MD & GCN & $0.285 \pm 0.065$ & $0.204 \pm 0.127$ & $0.865 \pm 0.382$ & 6.9 \\
MUTAG & GCN & $0.162 \pm 0.045$ & $0.204 \pm 0.046$ & $0.494 \pm 0.114$ & 51.2 \\
Mutagenicity & GCN & $0.153 \pm 0.012$ & $0.052 \pm 0.012$ & $0.494 \pm 0.053$ & 86.8 \\
\bottomrule
\end{tabular}

\end{table*}

Held-out GCN AUROC ranged from $0.561 \pm 0.133$ on COX2\_MD to $0.985 \pm 0.015$ on AIDS. The prespecified weak-predictor rule flagged COX2\_MD; its explanation results characterize that frozen model rather than application utility.

\subsection{Semantic coverage and conditional quality (RQ1--RQ2)}

Table~\ref{tab:all-dataset-explanations} reports all three proposed
explanations for every dataset. Success uses all attempted
graph--model--explainer units as its denominator; the remaining quantities are
conditional on success and are therefore never interpreted without coverage.

\begin{table*}[t]
\centering
\caption{Primary GCN multi-perspective explanations on all five
datasets. Success is the percentage of all attempts; continuous outcomes are mean
$\pm$ standard deviation after averaging explainer repetitions within each
graph--model pair. Comp. denotes comprehensiveness and $\Delta p_y$ is the
drop in the original-class probability.}
\label{tab:all-dataset-explanations}
\fontsize{6.5}{8}\selectfont
\setlength{\tabcolsep}{3pt}
\begin{tabular}{lccc|ccc|ccc}
\toprule
& \multicolumn{3}{c|}{Factual} & \multicolumn{3}{c|}{Counterfactual} & \multicolumn{3}{c}{Exemplar} \\
Dataset & Success & Retained & Comp. & Success & Edits & $\Delta p_y$ & Success & Edited & Retained $p_y$ ratio \\
\midrule
MUTAG & 100.0\% & $0.105 \pm 0.131$ & $0.120 \pm 0.220$ & 67.6\% & $2.59 \pm 1.57$ & $0.436 \pm 0.192$ & 98.9\% & $0.279 \pm 0.048$ & $1.041 \pm 0.161$ \\
Mutagenicity & 100.0\% & $0.155 \pm 0.203$ & $0.142 \pm 0.241$ & 63.7\% & $2.85 \pm 1.91$ & $0.315 \pm 0.155$ & 99.0\% & $0.246 \pm 0.068$ & $1.046 \pm 0.129$ \\
AIDS & 100.0\% & $0.103 \pm 0.079$ & $0.019 \pm 0.113$ & 4.8\% & $3.48 \pm 2.10$ & $0.419 \pm 0.165$ & 100.0\% & $0.285 \pm 0.059$ & $1.002 \pm 0.040$ \\
COX2\_MD & 100.0\% & $0.092 \pm 0.218$ & $0.048 \pm 0.173$ & 5.5\% & $4.68 \pm 2.33$ & $0.099 \pm 0.114$ & 100.0\% & $0.025 \pm 0.005$ & $0.998 \pm 0.028$ \\
BBBP & 100.0\% & $0.086 \pm 0.093$ & $0.090 \pm 0.163$ & 54.6\% & $3.65 \pm 2.24$ & $0.382 \pm 0.146$ & 99.7\% & $0.277 \pm 0.052$ & $1.008 \pm 0.094$ \\
\bottomrule
\end{tabular}

\end{table*}

Across datasets, hard factual coverage was 100.0\% and successful factuals retained 0.09--0.16 of input edges. Counterfactual flip coverage was 4.8--67.6\%; successful flips removed 2.59--4.68 edges on average, depending on the dataset. Exemplar preservation coverage was 98.9--100.0\%; successful exemplars edited 0.02--0.29 of edges while retaining 1.00--1.05 of the query-class confidence. These conditional ranges are interpreted only together with their coverage denominators.

Factual coverage must be read in light of the candidate family: its full-graph
prefix reproduces the query prediction. Complete coverage is therefore
expected, not evidence that the learned ranking is uniquely informative.
Compactness and comprehensiveness characterize the smaller returned mask.
Counterfactual and exemplar searches instead require a positive, bounded edit
and may have no successful candidate. High exemplar coverage consequently
indicates prediction-preserving edits in this space, not general model
robustness or stable attribution.

\paragraph{Interpreting factual sparsity.}
The factual projection deliberately scans retained-edge prefixes from one edge
upward and accepts the first mask that preserves the label and at least
$0.8p_y(G)$. Validation selection then prefers the smallest retained fraction
among successful configurations. Very sparse factual masks are therefore an
expected consequence of the stated edge-sufficiency objective when node
features and self-loops dominate; they are not, by themselves, evidence of a
small connected mechanism. Table~\ref{tab:runtime-connectivity} reports
isolated nodes and components, including nodes absent from retained-edge
incidence. The enlarged panels keep these nodes visible in grey rather than
silently displaying an induced subgraph.

The method-appropriate factual, counterfactual, and exemplar native-output comparisons are
reported separately in Tables~\ref{tab:factual-baselines}--
\ref{tab:exemplar-baselines}; an empty method--view combination is never
manufactured by reinterpreting another algorithm's output. Because the
counterfactual discretizers differ, these native-output comparisons are
descriptive, not an attribution of a coverage difference to the learned loss.

\begin{table*}[t]
\centering
\caption{Factual native-output comparisons. Success includes failures;
quality is conditional on success and uses graph--model means $\pm$ standard
deviation after averaging explainer repetitions.}
\label{tab:factual-baselines}
\scriptsize
\begin{tabular}{lllcc}
\toprule
Dataset & Method & Success & Retained & Comp. \\
\midrule
MUTAG & MPGX-CF & 100.0\% & $0.105 \pm 0.131$ & $0.120 \pm 0.220$ \\
MUTAG & GNNExplainer & 100.0\% & $0.160 \pm 0.210$ & $0.103 \pm 0.213$ \\
MUTAG & PGExplainer & 100.0\% & $0.219 \pm 0.280$ & $0.112 \pm 0.243$ \\
MUTAG & CF$^2$ reimplementation & 96.8\% & $0.445 \pm 0.154$ & $0.519 \pm 0.355$ \\
MUTAG & Random & 100.0\% & $0.183 \pm 0.222$ & $0.114 \pm 0.220$ \\
\addlinespace
Mutagenicity & MPGX-CF & 100.0\% & $0.155 \pm 0.203$ & $0.142 \pm 0.241$ \\
Mutagenicity & GNNExplainer & 100.0\% & $0.209 \pm 0.262$ & $0.118 \pm 0.226$ \\
Mutagenicity & PGExplainer & 100.0\% & $0.266 \pm 0.331$ & $0.140 \pm 0.247$ \\
Mutagenicity & CF$^2$ reimplementation & 96.8\% & $0.579 \pm 0.153$ & $0.417 \pm 0.283$ \\
Mutagenicity & Random & 100.0\% & $0.229 \pm 0.281$ & $0.103 \pm 0.214$ \\
\addlinespace
AIDS & MPGX-CF & 100.0\% & $0.103 \pm 0.079$ & $0.019 \pm 0.113$ \\
AIDS & GNNExplainer & 100.0\% & $0.112 \pm 0.112$ & $0.018 \pm 0.109$ \\
AIDS & PGExplainer & 100.0\% & $0.111 \pm 0.115$ & $0.014 \pm 0.103$ \\
AIDS & CF$^2$ reimplementation & 61.7\% & $0.405 \pm 0.186$ & $0.094 \pm 0.212$ \\
AIDS & Random & 100.0\% & $0.113 \pm 0.118$ & $0.017 \pm 0.106$ \\
\addlinespace
COX2\_MD & MPGX-CF & 100.0\% & $0.092 \pm 0.218$ & $0.048 \pm 0.173$ \\
COX2\_MD & GNNExplainer & 100.0\% & $0.115 \pm 0.248$ & $0.057 \pm 0.178$ \\
COX2\_MD & PGExplainer & 100.0\% & $0.177 \pm 0.322$ & $0.072 \pm 0.206$ \\
COX2\_MD & CF$^2$ reimplementation & 56.5\% & $0.112 \pm 0.151$ & $0.139 \pm 0.264$ \\
COX2\_MD & Random & 100.0\% & $0.085 \pm 0.192$ & $0.031 \pm 0.130$ \\
\addlinespace
BBBP & MPGX-CF & 100.0\% & $0.086 \pm 0.093$ & $0.090 \pm 0.163$ \\
BBBP & GNNExplainer & 100.0\% & $0.145 \pm 0.169$ & $0.071 \pm 0.128$ \\
BBBP & PGExplainer & 100.0\% & $0.186 \pm 0.248$ & $0.072 \pm 0.141$ \\
BBBP & CF$^2$ reimplementation & 95.3\% & $0.542 \pm 0.122$ & $0.521 \pm 0.269$ \\
BBBP & Random & 100.0\% & $0.182 \pm 0.210$ & $0.067 \pm 0.132$ \\
\addlinespace
\bottomrule
\end{tabular}

\end{table*}

\begin{table*}[t]
\centering
\caption{Counterfactual native-output comparisons under the locked budget.
Success includes failures; quality is conditional on success. Native
discretization and optimization procedures differ across methods.}
\label{tab:counterfactual-baselines}
\scriptsize
\begin{tabular}{lllcc}
\toprule
Dataset & Method & Success & Edits & $\Delta p_y$ \\
\midrule
MUTAG & MPGX-CF & 67.6\% & $2.594 \pm 1.569$ & $0.436 \pm 0.192$ \\
MUTAG & CF-GNNExplainer & 3.7\% & $1.190 \pm 0.402$ & $0.398 \pm 0.118$ \\
MUTAG & CF$^2$ reimplementation & 7.2\% & $5.854 \pm 1.663$ & $0.756 \pm 0.202$ \\
MUTAG & Random & 31.3\% & $3.452 \pm 1.910$ & $0.396 \pm 0.187$ \\
\addlinespace
Mutagenicity & MPGX-CF & 63.7\% & $2.851 \pm 1.912$ & $0.315 \pm 0.155$ \\
Mutagenicity & CF-GNNExplainer & 2.2\% & $1.286 \pm 0.590$ & $0.335 \pm 0.127$ \\
Mutagenicity & CF$^2$ reimplementation & 1.3\% & $5.281 \pm 2.098$ & $0.321 \pm 0.209$ \\
Mutagenicity & Random & 22.5\% & $3.568 \pm 1.937$ & $0.231 \pm 0.140$ \\
\addlinespace
AIDS & MPGX-CF & 4.8\% & $3.475 \pm 2.100$ & $0.419 \pm 0.165$ \\
AIDS & CF-GNNExplainer & 0.1\% & $1.000 \pm 0.000$ & $0.380 \pm 0.090$ \\
AIDS & CF$^2$ reimplementation & 0.2\% & $4.300 \pm 1.925$ & $0.408 \pm 0.268$ \\
AIDS & Random & 1.2\% & $3.628 \pm 2.146$ & $0.330 \pm 0.173$ \\
\addlinespace
COX2\_MD & MPGX-CF & 5.5\% & $4.679 \pm 2.328$ & $0.099 \pm 0.114$ \\
COX2\_MD & CF-GNNExplainer & 0.0\% & -- & -- \\
COX2\_MD & CF$^2$ reimplementation & 0.3\% & $6.667 \pm 0.471$ & $0.041 \pm 0.052$ \\
COX2\_MD & Random & 1.0\% & $5.678 \pm 1.934$ & $0.060 \pm 0.059$ \\
\addlinespace
BBBP & MPGX-CF & 54.6\% & $3.647 \pm 2.244$ & $0.382 \pm 0.146$ \\
BBBP & CF-GNNExplainer & 0.2\% & $1.000 \pm 0.000$ & $0.369 \pm 0.000$ \\
BBBP & CF$^2$ reimplementation & 0.5\% & $6.250 \pm 1.255$ & $0.466 \pm 0.242$ \\
BBBP & Random & 10.6\% & $3.951 \pm 2.054$ & $0.238 \pm 0.131$ \\
\addlinespace
\bottomrule
\end{tabular}

\end{table*}

\begin{table*}[t]
\centering
\caption{Exemplar comparisons. Coverage requires a non-trivial
prediction-preserving edit; conditional quality uses the same aggregation
as Table~\ref{tab:all-dataset-explanations}.}
\label{tab:exemplar-baselines}
\scriptsize
\begin{tabular}{lllcc}
\toprule
Dataset & Method & Success & Edited & Retained $p_y$ ratio \\
\midrule
MUTAG & MPGX-CF & 98.9\% & $0.279 \pm 0.048$ & $1.041 \pm 0.161$ \\
MUTAG & Greedy control & 100.0\% & $0.305 \pm 0.026$ & $1.169 \pm 0.232$ \\
MUTAG & Random & 94.6\% & $0.268 \pm 0.060$ & $1.023 \pm 0.145$ \\
\addlinespace
Mutagenicity & MPGX-CF & 99.0\% & $0.246 \pm 0.068$ & $1.046 \pm 0.129$ \\
Mutagenicity & Greedy control & 99.9\% & $0.257 \pm 0.069$ & $1.191 \pm 0.211$ \\
Mutagenicity & Random & 96.5\% & $0.231 \pm 0.074$ & $1.027 \pm 0.127$ \\
\addlinespace
AIDS & MPGX-CF & 100.0\% & $0.285 \pm 0.059$ & $1.002 \pm 0.040$ \\
AIDS & Greedy control & 100.0\% & $0.287 \pm 0.058$ & $1.021 \pm 0.065$ \\
AIDS & Random & 99.7\% & $0.285 \pm 0.059$ & $1.003 \pm 0.040$ \\
\addlinespace
COX2\_MD & MPGX-CF & 100.0\% & $0.025 \pm 0.005$ & $0.998 \pm 0.028$ \\
COX2\_MD & Greedy control & 100.0\% & $0.025 \pm 0.005$ & $1.018 \pm 0.059$ \\
COX2\_MD & Random & 100.0\% & $0.025 \pm 0.005$ & $0.998 \pm 0.027$ \\
\addlinespace
BBBP & MPGX-CF & 99.7\% & $0.277 \pm 0.052$ & $1.008 \pm 0.094$ \\
BBBP & Greedy control & 100.0\% & $0.286 \pm 0.052$ & $1.137 \pm 0.186$ \\
BBBP & Random & 98.9\% & $0.268 \pm 0.059$ & $0.998 \pm 0.084$ \\
\addlinespace
\bottomrule
\end{tabular}

\end{table*}

The exemplar control is particularly informative: a simple greedy deletion
ranking matches or exceeds the proposed method's preservation coverage and
edited fraction on these datasets (Table~\ref{tab:exemplar-baselines}). The
three-view contribution is therefore a common semantic and audit protocol,
not evidence that the learned exemplar ranking is superior to this control.

\subsection{Complementarity and repeatability (RQ3--RQ4)}

Table~\ref{tab:cross-perspective-overlap} compares factual retained edges with
the deletions returned by the other views on complete successful triplets.
Mean factual--counterfactual overlap exceeds factual--exemplar overlap in
each dataset. This is consistent with distinct roles for decision-changing
and prediction-preserving deletions, but their differing sizes and the
successful-triplet restriction affect Jaccard values. It does not establish
independent explanations or distinct biochemical mechanisms.

Prediction preservation is not mask repeatability. Mean seed-level Jaccard stability ranges across datasets from 0.06--0.42 for factual retained sets, 0.31--0.94 for counterfactual deletions, and 0.02--0.24 for exemplar deletions. These stability values are conditional on having at least two successful repetitions, not on all attempted searches.

In particular, the low factual and exemplar repeatability means that multiple
edge sets can satisfy similar prediction constraints. These outputs are
alternative audits of one classifier, not uniquely recovered structural truth.
Conditional centrality changes and per-view runtime are reported in the
appendix; runtime includes failed searches and should be interpreted in the
recorded execution environment.

\subsection{Exploratory projection and representation controls (RQ4--RQ5)}

Table~\ref{tab:common-projection} re-evaluates stored counterfactual rankings
on the same 24 held-out graph IDs per dataset with three predictor seeds
($72$ graph--model units per dataset). Every method receives the same ranked
deletion-prefix scan and the same eight-edge and 35\%-of-edges caps. In this
controlled subset, the same learned scores can yield substantially different
coverage when their hardening procedure changes.
On the common-projection BBBP subset, CF$^2$ coverage changes from 2.8\% to 63.9\%, compared with 65.3\% for the proposed ranking. CF-GNNExplainer changes from 1.4\% to 54.2\%. On AIDS, CF$^2$ and the proposed ranking both reach 6.9\%. The native-output gap therefore cannot be assigned solely to the learned objective.

The proposed 0.5-threshold ablation and detailed protocol are reported in
Appendix~\ref{sec:reviewer-controls}.

The low counterfactual coverage on AIDS and COX2\_MD is not a universal
failure rate of graph counterfactual search: it is tied to this frozen
predictor and deletion budget.
On the same selected units, increasing the absolute cap from 8 to 32 while retaining the fractional cap changes COX2\_MD coverage from 9.7\% to 23.6\%; AIDS remains at 6.9\%. This is a budget sensitivity result, not a redefinition of the primary outcome.

COX2\_MD predictor AUROC is also weak, so these explanations
are useful for auditing that model, not for chemical conclusions.

The factual null further distinguishes edge sufficiency from a molecular
fragment.
The zero-edge input, which keeps node features and predictor self-loops, satisfies the factual label/confidence predicate for 48.6\%--91.7\% of selected graph--model units across datasets. With only nodes incident to the selected factual edges retained, sufficiency is 37.5\%--61.1\%. The latter intervention changes the model input and does not itself produce a validated molecular fragment.

These observations support the full-node interpretation of the mask backend.
Full values appear in
Appendix~\ref{sec:reviewer-controls}.

\begin{figure*}[t]
\centering
\includegraphics[width=\textwidth]{\artifactroot/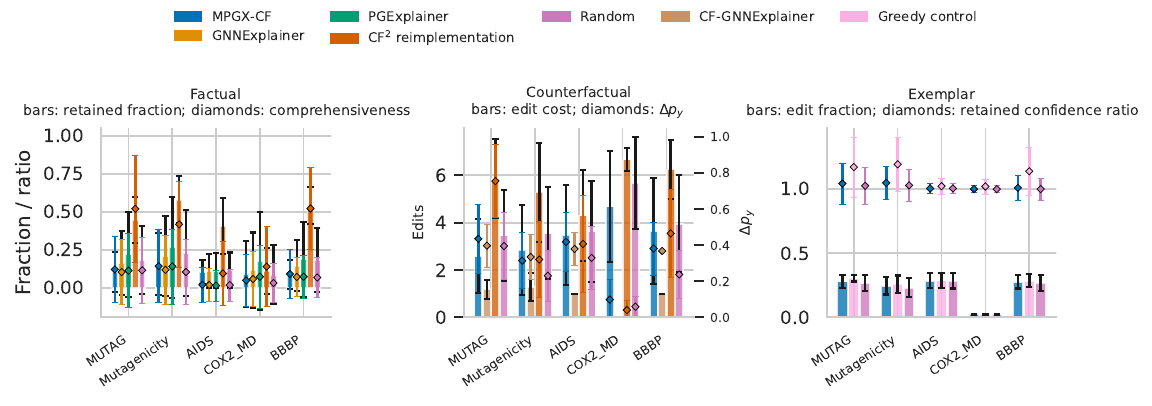}
\caption{Conditional quality of successful factual, counterfactual, and
exemplar explanations. Structural size and prediction behavior are shown
together because neither alone establishes semantic validity. Diamonds share
the fractional axis except in the counterfactual panel, where they use the
right-hand probability-drop axis.}
\label{fig:quality-bars}
\end{figure*}

Figure~\ref{fig:coverage-bars} presents semantic coverage before conditional
quality. Figure~\ref{fig:quality-bars} then reports the structural and
probabilistic outcomes of successful explanations with standard-deviation
error bars. This order prevents a method with low coverage from appearing
strong merely because its unsuccessful cases were removed.

\begin{figure*}[t]
\centering
\includegraphics[width=\textwidth]{\artifactroot/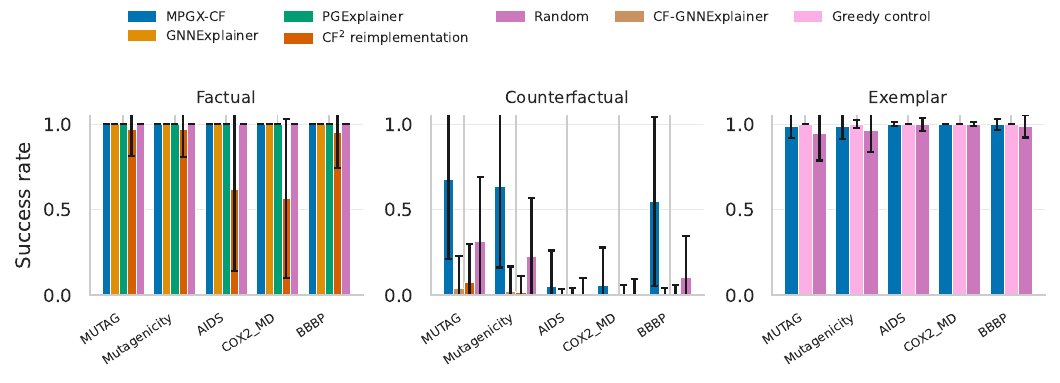}
\caption{Hard-constraint coverage for compatible methods on all five
datasets. Bars are means and error bars are standard deviations over
graph--model analysis units after averaging explainer repetitions.}
\label{fig:coverage-bars}
\end{figure*}

The appendix statistical tables contain conditional compactness,
stability, structural diagnostics, runtime, cross-perspective overlap, paired
effect estimates, and Holm-adjusted $p$ values. No analogous attribution
accuracy is reported for TU edge labels because they encode input categories,
not explanation annotations.

\subsection{Qualitative successes and failures}

\begin{figure*}[t]
\centering
\includegraphics[width=0.90\textwidth]{\artifactroot/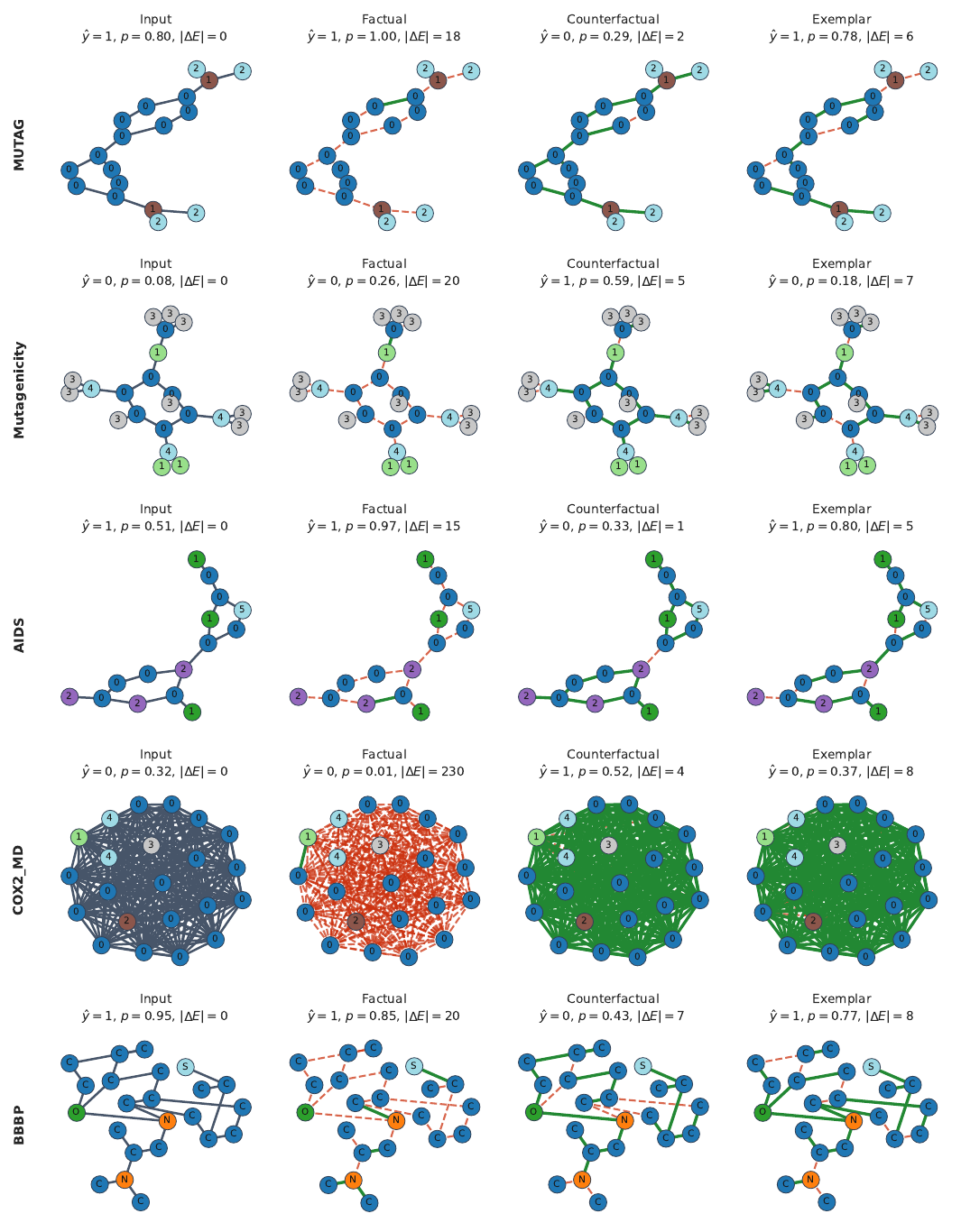}
\input{\artifactroot/submission_five_dataset_explanations_caption.tex}
\caption{Input, factual, counterfactual, and exemplar panels for MUTAG,
Mutagenicity, AIDS, COX2\_MD, and BBBP. A common graph and fixed layout are
used within each row; panels print class-1 probability, label, and deleted
unique-edge count. TU graphs are categorical renderings, not reconstructed
or sanitized molecules. Unavailable rows remain visible rather than being
silently omitted; the primary chemistry-aware BBBP case is shown separately
in Fig.~\ref{fig:bbbp-verified-fragments}. \overviewprovenance}
\label{fig:overview}
\end{figure*}

Figure~\ref{fig:overview} presents all three explanation views for all five
datasets. It is generated by the deterministic median
rule. Red dashed edges are excluded from the returned graph and retained edges
are green. When no complete successful triplet exists, the affected panel is
marked as a failure rather than removing the dataset row. The appendices also
provide one full-width strip per dataset in the visual style of the primary
BBBP case study, with the same locked cases and provenance.

\begin{figure*}[t]
\centering
\includegraphics[width=\textwidth]{\artifactroot/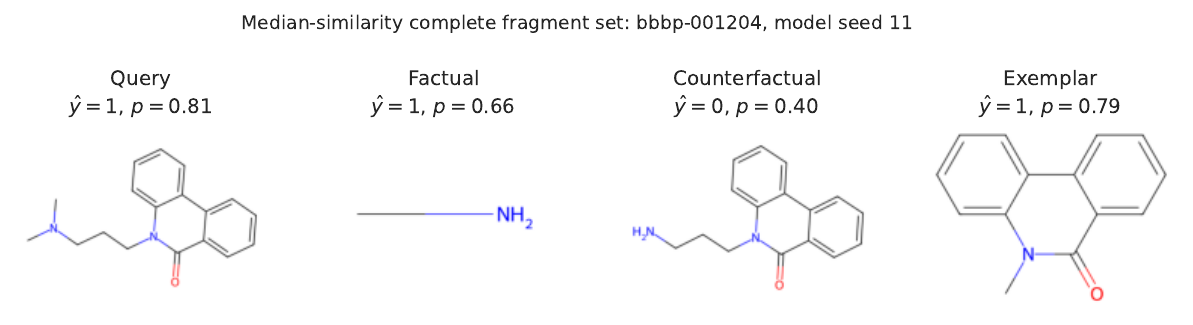}
\input{\artifactroot/submission_primary_bbbp_molecule_caption.tex}
\caption{Primary-test chemistry-aware BBBP fragment explanations. The factual and
exemplar fragments preserve the query decision, whereas the counterfactual
fragment reverses it. Probabilities printed in the panels are the frozen GCN's
class-1 probabilities. \bbbpprimaryprovenance}
\label{fig:bbbp-verified-fragments}
\end{figure*}

Figure~\ref{fig:bbbp-verified-fragments} reports a primary-test BBBP
chemistry-aware case. Unlike the edge-mask panels above, every returned object
is an RDKit-sanitized retained fragment. Its factual criterion minimizes
retained heavy atoms rather than retained edges. The case illustrates
the three semantic outcomes; population-level coverage and quality remain in
Table~\ref{tab:chemistry-fragments}. Sanitation makes this an analysis of
representable molecular variants, not experimental evidence that the predicted
property changes in a synthesized compound.

\FloatBarrier

\section{Limitations and Responsible Use}

\paragraph{Model and representation dependence.}
MPGE explains a trained classifier, not a causal chemical mechanism. A
factual mask can preserve a prediction through node features and self-messages
without identifying a compact molecular fragment. COX2\_MD also uses complete
distance graphs whose edge attributes are not consumed by the primary GCN.
Weak predictive performance, calibration errors, and altered-input
distributions limit chemical interpretation \cite{rao2022molecularxai,chen2023d4explainer}.

\paragraph{Search and evaluation scope.}
Ranked prefixes and fragment beams explore restricted deletion families;
success verifies a returned candidate, whereas failure is not proof of global
infeasibility. Seed-dependent masks can satisfy similar constraints, and
prediction preservation does not certify robust attribution
\cite{agarwal2022probing,li2024fragile}. Quality comparisons are conditional
on success, native methods have different discretizers and optimization
budgets, and the exploratory controls cover a selected subset. The primary
study has three GCN seeds, eight completed AIDS explanation folds, and one
BBBP scaffold split. GINE validation checks do not establish held-out
architectural robustness or performance on new training samples.

\paragraph{Chemical and human validity.}
RDKit sanitation establishes representability, not synthesizability,
stability, safety, or experimentally changed activity \cite{rdkitbook}.
The fragment backend cannot search atom or bond additions and does not enforce
a learned molecular distribution. Expert-chemist evaluation and experimental
property validation remain necessary before claiming medicinal-chemistry
utility \cite{rao2022molecularxai}. The figures support model inspection and
hypothesis generation, not chemical intervention recommendations.

\paragraph{Independent reproducibility.}
The upload provides generated artifacts, metric summaries, and compact claim
and case provenance. The absence of a public row-level record and checkpoint
accession still limits independent reruns; local completeness should not be
equated with a public reproducibility release.

\section{Conclusion}

\framework{} connects three questions about one molecular graph prediction:
what information supports it, what bounded deletion changes it, and what
non-trivial deletion preserves it. View-specific objectives and hard-output
checks make these questions operational in a graph-classification extension
of CF-GNNExplainer, with a separate sanitized-fragment instantiation for BBBP.
The evaluated GCNs yielded 4.8--67.6\% bounded counterfactual coverage and 98.9--100.0\% non-trivial exemplar coverage across the five datasets, illustrating why semantic coverage and conditional quality must be reported separately.

Factual compactness is informative only under the documented full-node
mask semantics, and high preservation coverage does not imply stable masks.
The exploratory controls further show that discretization, retained features,
and edit budgets materially shape the evidence. The resulting framework
supports transparent multi-perspective model auditing; establishing broader
architectural, chemical, or human utility remains a separate validation task.

\clearpage
\onecolumn
\appendices
\section{Chemistry-Aware BBBP Fragment Analysis}

The fragment backend cuts only non-ring single bonds, canonicalizes each
retained component, restores implicit hydrogens, and accepts only candidates
that pass RDKit sanitization. Its search space and edit unit differ from the
edge-mask backend, so these results are reported separately rather than pooled
with the structural comparison. Failures remain in the coverage denominator;
conditional rows summarize only returned sanitized molecules. Tanimoto
similarity uses radius-two Morgan fingerprints, and retained confidence is
relative to the unmodified query.

The primary-test qualitative fragment triplet appears once, in
Fig.~\ref{fig:bbbp-verified-fragments}. It is kept separate from the
structural edge-mask strips because sanitization, fragment retention, and atom
removal define a different intervention space.

\begin{table}[H]
\centering
\caption{GCN chemistry-aware fragment search on the scaffold-held-out BBBP
test set. Success is percentage-only over all attempts, including failures;
other values are conditional means $\pm$ standard deviation over graph--model
units. Fragment factual size is retained heavy atoms, not retained edges.}
\label{tab:chemistry-fragments}
\scriptsize
\begin{tabular}{lcccccc}
\toprule
View & Success & Steps & Atoms removed & Tanimoto & Retained $p_y$ ratio & Candidates \\
\midrule
Factual & 97.7\% & $1.71 \pm 0.70$ & $19.91 \pm 8.79$ & $0.10 \pm 0.11$ & $0.92 \pm 0.12$ & $56.98 \pm 59.75$ \\
Counterfactual & 27.4\% & $1.40 \pm 0.61$ & $3.22 \pm 2.66$ & $0.62 \pm 0.14$ & $0.56 \pm 0.15$ & $73.96 \pm 66.97$ \\
Exemplar & 96.4\% & $2.19 \pm 0.78$ & $6.63 \pm 2.60$ & $0.35 \pm 0.12$ & $1.00 \pm 0.14$ & $57.54 \pm 59.95$ \\
\bottomrule
\end{tabular}

\end{table}

\FloatBarrier
\section{Validation-Only Selection}

The complete $3\times3$ perspective-specific regularization grid is evaluated independently for each
perspective on deterministic class-balanced validation subsets. The GCN grid
sets the primary test configuration; GINE validation results are displayed as
an architectural validation sensitivity check, not a completed GINE
held-out explanation comparison. The factual and counterfactual losses use
size/quadratic penalties; the exemplar uses an edit reward and entropy penalty.
Test graphs never participate in selection.

\begin{figure}[H]
\centering
\includegraphics[width=\textwidth]{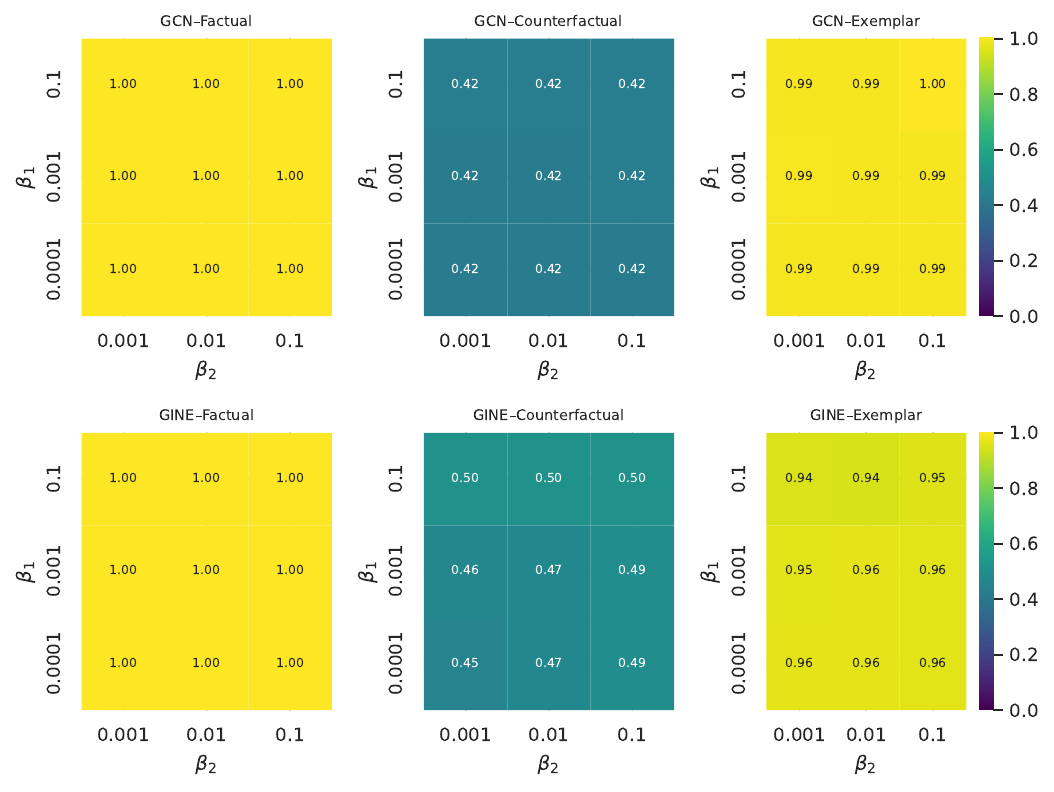}
\caption{Macro-averaged hard-constraint success over the complete GCN and GINE
validation grids. Failed searches remain in every denominator.}
\end{figure}

\begin{figure}[H]
\centering
\includegraphics[width=\textwidth]{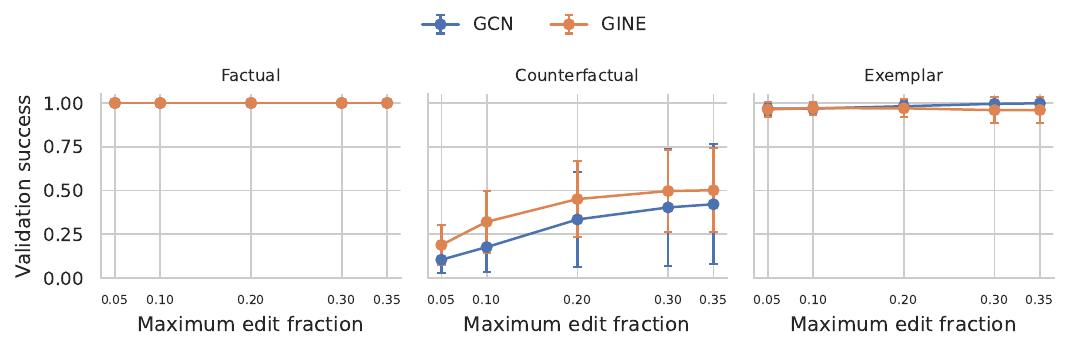}
\caption{Validation coverage as the permitted edit fraction changes. Error
bars show variation across dataset-level means.}
\end{figure}

\FloatBarrier
\section{Exploratory Held-Out Controls}
\label{sec:reviewer-controls}

The control population contains 24 graph IDs per dataset, selected with fixed
seed 20261007 by predicted-class balance from complete counterfactual-ranking
records. All selected graphs belong to their outer-fold test set or BBBP
scaffold test set. Three frozen GCN checkpoints and one explainer seed (101)
yield 72 graph--model units per dataset. The common projection scans deletion
prefixes of stored edge rankings and evaluates every hard candidate with the
same checkpoint. This controls the discretization rule but does not equate
optimization budgets, objectives, or training procedures. It is exploratory
and is not used to revise validation-selected hyperparameters.

\begin{table}[H]
\centering
\caption{Exploratory counterfactual control on a class-balanced held-out
subset: native hard output versus identical ranked-prefix projection of each
method's stored scores. Percentages include failures; 72 graph--model units
represent 24 graph IDs and three predictor seeds per dataset.}
\label{tab:common-projection}
\scriptsize
\begin{tabular}{llrcc}
\toprule
Dataset & Ranking & Units & Native & Common projection \\
\midrule
MUTAG & MPGX-CF & 72 & 54.2\% & 54.2\% \\
MUTAG & CF-GNNExplainer & 72 & 5.6\% & 41.7\% \\
MUTAG & CF$^2$ & 72 & 4.2\% & 51.4\% \\
MUTAG & Random & 72 & 26.4\% & 26.4\% \\
\addlinespace
Mutagenicity & MPGX-CF & 72 & 63.9\% & 63.9\% \\
Mutagenicity & CF-GNNExplainer & 72 & 1.4\% & 51.4\% \\
Mutagenicity & CF$^2$ & 72 & 2.8\% & 59.7\% \\
Mutagenicity & Random & 72 & 19.4\% & 19.4\% \\
\addlinespace
AIDS & MPGX-CF & 72 & 6.9\% & 6.9\% \\
AIDS & CF-GNNExplainer & 72 & 0.0\% & 5.6\% \\
AIDS & CF$^2$ & 72 & 0.0\% & 6.9\% \\
AIDS & Random & 72 & 1.4\% & 1.4\% \\
\addlinespace
COX2\_MD & MPGX-CF & 72 & 9.7\% & 9.7\% \\
COX2\_MD & CF-GNNExplainer & 72 & 0.0\% & 8.3\% \\
COX2\_MD & CF$^2$ & 72 & 1.4\% & 6.9\% \\
COX2\_MD & Random & 72 & 1.4\% & 1.4\% \\
\addlinespace
BBBP & MPGX-CF & 72 & 65.3\% & 65.3\% \\
BBBP & CF-GNNExplainer & 72 & 1.4\% & 54.2\% \\
BBBP & CF$^2$ & 72 & 2.8\% & 63.9\% \\
BBBP & Random & 72 & 11.1\% & 11.1\% \\
\addlinespace
\bottomrule
\end{tabular}

\end{table}

The proposed soft mask itself flips far less often if a fixed 0.5 threshold is
used in place of ranked-prefix search (Table~\ref{tab:threshold-ablation}).
Thus the search/projection mechanism is a major part of the implemented
method, not a negligible post-processing detail.

\begin{table}[H]
\centering
\caption{Proposed counterfactual ranking: fixed threshold versus hard-prefix
search on the same selected held-out units.}
\label{tab:threshold-ablation}
\scriptsize
\begin{tabular}{lrcc}
\toprule
Dataset & Units & Proposed 0.5 threshold & Proposed ranked prefix \\
\midrule
MUTAG & 72 & 16.7\% & 54.2\% \\
Mutagenicity & 72 & 9.7\% & 63.9\% \\
AIDS & 72 & 1.4\% & 6.9\% \\
COX2\_MD & 72 & 0.0\% & 9.7\% \\
BBBP & 72 & 15.3\% & 65.3\% \\
\bottomrule
\end{tabular}

\end{table}

\begin{table}[H]
\centering
\caption{Factual null and induced-node controls. The no-edge input retains all
query nodes, features, and predictor self-loops; the induced input retains
only nodes incident to the proposed factual edges. Percentages use all 72
graph--model units per dataset.}
\label{tab:factual-controls}
\scriptsize
\begin{tabular}{lrrcc}
\toprule
Dataset & Graphs & Units & No-edge feasible & Induced factual feasible \\
\midrule
MUTAG & 24 & 72 & 73.6\% & 37.5\% \\
Mutagenicity & 24 & 72 & 56.9\% & 55.6\% \\
AIDS & 24 & 72 & 91.7\% & 61.1\% \\
COX2\_MD & 24 & 72 & 69.4\% & 56.9\% \\
BBBP & 24 & 72 & 48.6\% & 38.9\% \\
\bottomrule
\end{tabular}

\end{table}

The eight-edge hard cap allows a small edit fraction in dense COX2\_MD graphs.
Table~\ref{tab:budget-cap} keeps the same learned rankings and 35\% relative
budget while changing only the absolute cap from 8 to 32. Coverage increases
on COX2\_MD, whereas AIDS remains unchanged. The latter observation alone
does not identify which model or search constraint prevents additional flips.
This is a sensitivity analysis, not a redefinition of the
locked primary outcome. The full row-level records, selected graph IDs, and
selection rule are in \texttt{reviewer\_controls}.

\begin{table}[H]
\centering
\caption{Exploratory hard-budget sensitivity under a common projection on
the same class-balanced held-out subset.}
\label{tab:budget-cap}
\scriptsize
\begin{tabular}{llrcc}
\toprule
Dataset & Ranking & Units & Cap 8 & Cap 32 \\
\midrule
AIDS & MPGX-CF & 72 & 6.9\% & 6.9\% \\
AIDS & CF-GNNExplainer & 72 & 5.6\% & 5.6\% \\
AIDS & CF$^2$ & 72 & 6.9\% & 6.9\% \\
AIDS & Random & 72 & 1.4\% & 1.4\% \\
\addlinespace
COX2\_MD & MPGX-CF & 72 & 9.7\% & 23.6\% \\
COX2\_MD & CF-GNNExplainer & 72 & 8.3\% & 22.2\% \\
COX2\_MD & CF$^2$ & 72 & 6.9\% & 22.2\% \\
COX2\_MD & Random & 72 & 1.4\% & 5.6\% \\
\addlinespace
\bottomrule
\end{tabular}

\end{table}

\FloatBarrier
\section{Statistical Tables}

For each continuous metric, explainer repetitions are first averaged within a
graph--model pair. Descriptive intervals use 10,000 hierarchical bootstrap
resamples. For paired method effects, repeated predictor-seed differences are
averaged within each held-out graph before a graph-level bootstrap and
Wilcoxon signed-rank test. Thus a graph, not a graph--model pair, is the
resampling and test unit; shared fitted models can still induce dependence.
Holm adjustment is performed within each
dataset--architecture--perspective--outcome family. The local analysis also records
graph count, model-seed count, and failure denominator. Dataset domains are
never micro-averaged. The machine-readable sources are
\texttt{model\_summary.parquet}, \texttt{explanation\_summary.parquet}, and
\texttt{paired\_method\_comparisons.parquet}. The upload's
\texttt{manuscript\_metric\_summary.csv} includes estimates, descriptive
intervals, observation counts, analysis-unit counts, and success conditioning.
Hierarchical interval centers weight folds and predictor seeds equally, while
descriptive means weight available graph--model units equally; these
estimands need not coincide when fold sizes or successful populations differ.

\begin{table}[H]
\centering
\caption{Proposed GCN per-view runtime and connectivity. Runtime is
unconditional wall time in seconds, including failed searches; components
and isolated nodes are conditional on hard success, with predictor self-loops
excluded. Values are mean $\pm$ standard deviation after averaging explainer
seeds within graph--model units. Units counts timing units, not independent
training replications. Runtime intervals are descriptive hierarchical 95\%
bootstrap intervals; hardware and implementation limit timing comparisons.}
\label{tab:runtime-connectivity}
\small
\begin{tabular}{llccccc}
\toprule
Dataset & View & Runtime (s) & Runtime 95\% CI & Components & Isolated nodes & Units \\
\midrule
MUTAG & Factual & $0.54 \pm 0.04$ & $[0.53, 0.56]$ & $15.99 \pm 4.95$ & $14.64 \pm 5.12$ & 564 \\
MUTAG & Counterfactual & $0.55 \pm 0.04$ & $[0.54, 0.55]$ & $2.00 \pm 1.00$ & $0.52 \pm 0.79$ & 564 \\
MUTAG & Exemplar & $0.57 \pm 0.04$ & $[0.56, 0.58]$ & $4.02 \pm 1.08$ & $1.46 \pm 0.78$ & 564 \\
\addlinespace
Mutagenicity & Factual & $0.58 \pm 0.22$ & $[0.51, 0.68]$ & $26.02 \pm 19.56$ & $23.52 \pm 19.58$ & 13011 \\
Mutagenicity & Counterfactual & $0.56 \pm 0.16$ & $[0.51, 0.64]$ & $3.57 \pm 6.14$ & $1.79 \pm 6.04$ & 13011 \\
Mutagenicity & Exemplar & $0.58 \pm 0.16$ & $[0.52, 0.65]$ & $6.79 \pm 9.10$ & $3.92 \pm 8.99$ & 13011 \\
\addlinespace
AIDS & Factual & $0.55 \pm 0.17$ & $[0.49, 0.64]$ & $14.45 \pm 13.69$ & $13.35 \pm 13.71$ & 4800 \\
AIDS & Counterfactual & $0.54 \pm 0.15$ & $[0.49, 0.62]$ & $3.31 \pm 1.64$ & $1.08 \pm 1.07$ & 4800 \\
AIDS & Exemplar & $0.56 \pm 0.14$ & $[0.50, 0.63]$ & $3.79 \pm 1.54$ & $1.51 \pm 0.99$ & 4800 \\
\addlinespace
COX2\_MD & Factual & $0.63 \pm 0.16$ & $[0.58, 0.69]$ & $19.78 \pm 9.64$ & $18.61 \pm 9.82$ & 909 \\
COX2\_MD & Counterfactual & $0.59 \pm 0.07$ & $[0.56, 0.63]$ & $1.00 \pm 0.00$ & $0.00 \pm 0.00$ & 909 \\
COX2\_MD & Exemplar & $0.61 \pm 0.07$ & $[0.58, 0.65]$ & $1.00 \pm 0.00$ & $0.00 \pm 0.00$ & 909 \\
\addlinespace
BBBP & Factual & $0.50 \pm 0.03$ & $[0.49, 0.50]$ & $23.35 \pm 8.73$ & $21.63 \pm 8.94$ & 609 \\
BBBP & Counterfactual & $0.50 \pm 0.03$ & $[0.50, 0.51]$ & $3.31 \pm 1.51$ & $1.24 \pm 1.11$ & 609 \\
BBBP & Exemplar & $0.52 \pm 0.03$ & $[0.51, 0.53]$ & $5.57 \pm 1.04$ & $2.07 \pm 0.86$ & 609 \\
\addlinespace
\bottomrule
\end{tabular}

\end{table}

\begin{table}[H]
\centering
\caption{Structural and probability diagnostics for the proposed method.
Continuous metrics are conditional on hard success; stability is computed
across explainer seeds within the same graph--model unit.}
\label{tab:structural-diagnostics}
\tiny
\begin{tabular}{llccccccc}
\toprule
Dataset & View & Present & Removed & Removed frac. & $|\Delta p|$ & $\Delta$ betw. & $\Delta$ harmonic & Stability \\
\midrule
MUTAG & Factual & $1.98 \pm 2.71$ & $17.82 \pm 6.05$ & $0.90 \pm 0.13$ & $0.08 \pm 0.09$ & $0.22 \pm 0.04$ & $6.27 \pm 1.25$ & $0.25 \pm 0.35$ \\
MUTAG & Counterfactual & $17.33 \pm 5.22$ & $2.59 \pm 1.57$ & $0.13 \pm 0.07$ & $0.44 \pm 0.19$ & $0.12 \pm 0.05$ & $1.68 \pm 0.97$ & $0.74 \pm 0.29$ \\
MUTAG & Exemplar & $14.27 \pm 4.33$ & $5.52 \pm 1.75$ & $0.28 \pm 0.05$ & $0.08 \pm 0.08$ & $0.16 \pm 0.04$ & $3.33 \pm 0.78$ & $0.19 \pm 0.13$ \\
\addlinespace
Mutagenicity & Factual & $4.33 \pm 6.97$ & $26.44 \pm 16.25$ & $0.84 \pm 0.20$ & $0.12 \pm 0.10$ & $0.21 \pm 0.05$ & $7.92 \pm 2.50$ & $0.42 \pm 0.38$ \\
Mutagenicity & Counterfactual & $26.13 \pm 15.08$ & $2.85 \pm 1.91$ & $0.11 \pm 0.07$ & $0.31 \pm 0.16$ & $0.10 \pm 0.06$ & $2.44 \pm 1.28$ & $0.74 \pm 0.30$ \\
Mutagenicity & Exemplar & $24.11 \pm 15.79$ & $6.69 \pm 1.83$ & $0.25 \pm 0.07$ & $0.08 \pm 0.06$ & $0.16 \pm 0.06$ & $4.33 \pm 0.98$ & $0.17 \pm 0.13$ \\
\addlinespace
AIDS & Factual & $1.24 \pm 1.35$ & $14.96 \pm 15.03$ & $0.90 \pm 0.08$ & $0.02 \pm 0.05$ & $0.27 \pm 0.07$ & $5.13 \pm 1.88$ & $0.06 \pm 0.19$ \\
AIDS & Counterfactual & $14.45 \pm 4.53$ & $3.48 \pm 2.10$ & $0.19 \pm 0.09$ & $0.42 \pm 0.16$ & $0.17 \pm 0.07$ & $2.57 \pm 1.24$ & $0.94 \pm 0.14$ \\
AIDS & Exemplar & $12.23 \pm 13.45$ & $3.97 \pm 1.98$ & $0.29 \pm 0.06$ & $0.01 \pm 0.03$ & $0.20 \pm 0.07$ & $2.71 \pm 0.87$ & $0.24 \pm 0.10$ \\
\addlinespace
COX2\_MD & Factual & $30.99 \pm 75.25$ & $304.14 \pm 96.02$ & $0.91 \pm 0.22$ & $0.07 \pm 0.07$ & $0.01 \pm 0.03$ & $21.93 \pm 6.98$ & $0.08 \pm 0.22$ \\
COX2\_MD & Counterfactual & $313.68 \pm 51.22$ & $4.68 \pm 2.33$ & $0.02 \pm 0.01$ & $0.10 \pm 0.11$ & $0.00 \pm 0.00$ & $0.44 \pm 0.19$ & $0.31 \pm 0.28$ \\
COX2\_MD & Exemplar & $327.15 \pm 65.22$ & $7.97 \pm 0.34$ & $0.02 \pm 0.00$ & $0.01 \pm 0.02$ & $0.00 \pm 0.00$ & $0.47 \pm 0.04$ & $0.02 \pm 0.01$ \\
\addlinespace
BBBP & Factual & $2.19 \pm 2.35$ & $25.70 \pm 9.64$ & $0.91 \pm 0.09$ & $0.08 \pm 0.07$ & $0.23 \pm 0.03$ & $7.30 \pm 1.31$ & $0.38 \pm 0.35$ \\
BBBP & Counterfactual & $23.63 \pm 9.02$ & $3.65 \pm 2.24$ & $0.14 \pm 0.08$ & $0.38 \pm 0.15$ & $0.14 \pm 0.06$ & $2.61 \pm 1.22$ & $0.86 \pm 0.21$ \\
BBBP & Exemplar & $20.54 \pm 8.85$ & $7.35 \pm 1.19$ & $0.28 \pm 0.05$ & $0.05 \pm 0.05$ & $0.19 \pm 0.04$ & $4.06 \pm 0.56$ & $0.16 \pm 0.06$ \\
\addlinespace
\bottomrule
\end{tabular}

\end{table}

\begin{table}[H]
\centering
\caption{Cross-perspective overlap for the proposed method. Overlap is the
Jaccard similarity between factual retained edges and the edges deleted by
the counterfactual (CF) or exemplar view. Contrast is CF minus exemplar
overlap. Values are mean $\pm$ standard deviation after averaging explainer
seeds within graph--model units on complete successful triplets. Counts and
intervals are in the supplementary metric records.}
\label{tab:cross-perspective-overlap}
\scriptsize
\begin{tabular}{lccc}
\toprule
Dataset & Factual--CF overlap & Factual--exemplar overlap & Contrast \\
\midrule
MUTAG & $0.095 \pm 0.187$ & $0.017 \pm 0.048$ & $0.078 \pm 0.177$ \\
Mutagenicity & $0.150 \pm 0.212$ & $0.036 \pm 0.065$ & $0.114 \pm 0.214$ \\
AIDS & $0.259 \pm 0.211$ & $0.075 \pm 0.083$ & $0.184 \pm 0.204$ \\
COX2\_MD & $0.014 \pm 0.029$ & $0.005 \pm 0.008$ & $0.009 \pm 0.030$ \\
BBBP & $0.177 \pm 0.210$ & $0.026 \pm 0.048$ & $0.151 \pm 0.199$ \\
\bottomrule
\end{tabular}

\end{table}

The following tables report paired proposed-minus-baseline effects. Explainer
seeds and predictor seeds are averaged within graph before inference; brackets are graph-level 95\% bootstrap
intervals and multiplicity is controlled by Holm adjustment within each
dataset, architecture, perspective, and outcome family.
Coverage effects use every graph. Quality effects require both methods to
return a valid explanation and can therefore have very small $n$ when a
baseline rarely succeeds; these conditional effects are not population-wide
quality comparisons. Values printed as $p<10^{-300}$ reflect numerical
underflow, not an exactly zero probability.

\begin{table}[H]
\centering
\caption{Paired factual effects against compatible baselines.}
\tiny
\begin{tabular}{llcccc}
\toprule
Dataset & Baseline & Coverage difference & Holm $p$ & Comprehensiveness difference & Holm $p$ \\
\midrule
MUTAG & GNNExplainer & $0.000$ [0.000, 0.000]; $n=188$ & 1 & $0.018$ [0.006, 0.029]; $n=188$ & 0.00729 \\
MUTAG & PGExplainer & $0.000$ [0.000, 0.000]; $n=188$ & 1 & $0.008$ [-0.008, 0.024]; $n=188$ & 1 \\
MUTAG & CF$^2$ reimplementation & $0.032$ [0.018, 0.048]; $n=188$ & 0.000139 & $-0.400$ [-0.445, -0.357]; $n=188$ & 8.57e-31 \\
MUTAG & Random & $0.000$ [0.000, 0.000]; $n=188$ & 1 & $0.006$ [-0.005, 0.018]; $n=188$ & 1 \\
\addlinespace
Mutagenicity & GNNExplainer & $0.000$ [0.000, 0.000]; $n=4337$ & 1 & $0.024$ [0.022, 0.026]; $n=4337$ & 4.79e-69 \\
Mutagenicity & PGExplainer & $0.000$ [0.000, 0.000]; $n=4337$ & 1 & $0.002$ [-0.001, 0.005]; $n=4337$ & 0.456 \\
Mutagenicity & CF$^2$ reimplementation & $0.032$ [0.028, 0.035]; $n=4337$ & 4.49e-72 & $-0.275$ [-0.282, -0.268]; $n=4314$ & $<10^{-300}$ \\
Mutagenicity & Random & $0.000$ [0.000, 0.000]; $n=4337$ & 1 & $0.038$ [0.036, 0.041]; $n=4337$ & 4.31e-150 \\
\addlinespace
AIDS & GNNExplainer & $0.000$ [0.000, 0.000]; $n=1600$ & 1 & $0.001$ [-0.000, 0.002]; $n=1600$ & 0.064 \\
AIDS & PGExplainer & $0.000$ [0.000, 0.000]; $n=1600$ & 1 & $0.005$ [0.003, 0.006]; $n=1600$ & 9.31e-19 \\
AIDS & CF$^2$ reimplementation & $0.383$ [0.362, 0.405]; $n=1600$ & 4.02e-139 & $-0.060$ [-0.068, -0.053]; $n=1164$ & 1.55e-163 \\
AIDS & Random & $0.000$ [0.000, 0.000]; $n=1600$ & 1 & $0.002$ [0.001, 0.003]; $n=1600$ & 8.61e-06 \\
\addlinespace
COX2\_MD & GNNExplainer & $0.000$ [0.000, 0.000]; $n=303$ & 1 & $-0.010$ [-0.019, -0.001]; $n=303$ & 0.00577 \\
COX2\_MD & PGExplainer & $0.000$ [0.000, 0.000]; $n=303$ & 1 & $-0.024$ [-0.034, -0.015]; $n=303$ & 8.05e-07 \\
COX2\_MD & CF$^2$ reimplementation & $0.435$ [0.407, 0.462]; $n=303$ & 2.41e-46 & $-0.125$ [-0.150, -0.101]; $n=297$ & 3.99e-42 \\
COX2\_MD & Random & $0.000$ [0.000, 0.000]; $n=303$ & 1 & $0.017$ [0.010, 0.025]; $n=303$ & 0.00577 \\
\addlinespace
BBBP & GNNExplainer & $0.000$ [0.000, 0.000]; $n=203$ & 1 & $0.020$ [0.012, 0.028]; $n=203$ & 0.000235 \\
BBBP & PGExplainer & $0.000$ [0.000, 0.000]; $n=203$ & 1 & $0.018$ [0.010, 0.026]; $n=203$ & 0.000104 \\
BBBP & CF$^2$ reimplementation & $0.047$ [0.025, 0.073]; $n=203$ & 0.000659 & $-0.426$ [-0.456, -0.397]; $n=199$ & 8.77e-34 \\
BBBP & Random & $0.000$ [0.000, 0.000]; $n=203$ & 1 & $0.023$ [0.015, 0.032]; $n=203$ & 2.32e-05 \\
\addlinespace
\bottomrule
\end{tabular}

\end{table}

\begin{table}[H]
\centering
\caption{Paired counterfactual effects against compatible baselines.}
\tiny
\begin{tabular}{llcccc}
\toprule
Dataset & Baseline & Coverage difference & Holm $p$ & Removed-edge difference & Holm $p$ \\
\midrule
MUTAG & CF-GNNExplainer & $0.638$ [0.580, 0.694]; $n=188$ & 2.08e-26 & $-0.141$ [-0.392, 0.088]; $n=17$ & 0.256 \\
MUTAG & CF$^2$ reimplementation & $0.603$ [0.546, 0.660]; $n=188$ & 8.94e-26 & $-3.370$ [-3.851, -2.876]; $n=47$ & 1.39e-08 \\
MUTAG & Random & $0.363$ [0.320, 0.406]; $n=188$ & 2.79e-24 & $-1.299$ [-1.530, -1.073]; $n=121$ & 2.45e-16 \\
\addlinespace
Mutagenicity & CF-GNNExplainer & $0.615$ [0.603, 0.628]; $n=4337$ & $<10^{-300}$ & $-0.130$ [-0.191, -0.070]; $n=239$ & 6.06e-06 \\
Mutagenicity & CF$^2$ reimplementation & $0.624$ [0.612, 0.637]; $n=4337$ & $<10^{-300}$ & $-3.581$ [-3.889, -3.275]; $n=184$ & 6.76e-29 \\
Mutagenicity & Random & $0.412$ [0.402, 0.423]; $n=4337$ & $<10^{-300}$ & $-1.694$ [-1.761, -1.626]; $n=1984$ & 8.45e-263 \\
\addlinespace
AIDS & CF-GNNExplainer & $0.046$ [0.038, 0.055]; $n=1600$ & 1.27e-21 & $0.000$ [0.000, 0.000]; $n=4$ & 1 \\
AIDS & CF$^2$ reimplementation & $0.045$ [0.037, 0.054]; $n=1600$ & 1.27e-21 & $-1.625$ [-2.250, -1.000]; $n=12$ & 0.00781 \\
AIDS & Random & $0.036$ [0.029, 0.043]; $n=1600$ & 1.57e-21 & $-1.380$ [-1.775, -1.003]; $n=51$ & 7.19e-07 \\
\addlinespace
COX2\_MD & CF-GNNExplainer & $0.055$ [0.040, 0.070]; $n=303$ & 4.76e-10 & -- & -- \\
COX2\_MD & CF$^2$ reimplementation & $0.051$ [0.038, 0.066]; $n=303$ & 6.26e-10 & $-2.917$ [-5.000, -0.833]; $n=4$ & 0.125 \\
COX2\_MD & Random & $0.045$ [0.032, 0.059]; $n=303$ & 8.95e-10 & $-2.742$ [-3.894, -1.712]; $n=11$ & 0.00391 \\
\addlinespace
BBBP & CF-GNNExplainer & $0.545$ [0.490, 0.600]; $n=203$ & 3.01e-26 & $0.000$ [0.000, 0.000]; $n=1$ & 1 \\
BBBP & CF$^2$ reimplementation & $0.541$ [0.486, 0.597]; $n=203$ & 3.33e-26 & $-3.500$ [-4.600, -2.100]; $n=5$ & 0.125 \\
BBBP & Random & $0.441$ [0.395, 0.487]; $n=203$ & 3.8e-26 & $-2.185$ [-2.591, -1.797]; $n=63$ & 1.04e-10 \\
\addlinespace
\bottomrule
\end{tabular}

\end{table}

\begin{table}[H]
\centering
\caption{Paired exemplar effects against compatible baselines.}
\tiny
\begin{tabular}{llcccc}
\toprule
Dataset & Baseline & Coverage difference & Holm $p$ & Edit-fraction difference & Holm $p$ \\
\midrule
MUTAG & Greedy control & $-0.011$ [-0.017, -0.005]; $n=188$ & 0.000941 & $-0.026$ [-0.030, -0.021]; $n=188$ & 1.6e-19 \\
MUTAG & Random & $0.044$ [0.030, 0.059]; $n=188$ & 1.35e-07 & $0.011$ [0.005, 0.017]; $n=188$ & 0.000736 \\
\addlinespace
Mutagenicity & Greedy control & $-0.010$ [-0.011, -0.008]; $n=4337$ & 1.15e-42 & $-0.011$ [-0.012, -0.010]; $n=4337$ & 9.02e-236 \\
Mutagenicity & Random & $0.025$ [0.022, 0.027]; $n=4337$ & 8.28e-74 & $0.015$ [0.014, 0.016]; $n=4335$ & 1.99e-157 \\
\addlinespace
AIDS & Greedy control & $-0.000$ [-0.001, 0.000]; $n=1600$ & 0.102 & $-0.002$ [-0.002, -0.001]; $n=1600$ & 7.09e-13 \\
AIDS & Random & $0.003$ [0.002, 0.004]; $n=1600$ & 0.000112 & $0.001$ [0.000, 0.001]; $n=1600$ & 0.0013 \\
\addlinespace
COX2\_MD & Greedy control & $0.000$ [0.000, 0.000]; $n=303$ & 1 & $-0.000$ [-0.000, 0.000]; $n=303$ & 0.0682 \\
COX2\_MD & Random & $0.000$ [0.000, 0.001]; $n=303$ & 0.635 & $0.000$ [-0.000, 0.000]; $n=303$ & 0.145 \\
\addlinespace
BBBP & Greedy control & $-0.003$ [-0.006, -0.001]; $n=203$ & 0.0198 & $-0.008$ [-0.011, -0.006]; $n=203$ & 8.74e-14 \\
BBBP & Random & $0.008$ [0.003, 0.014]; $n=203$ & 0.0198 & $0.010$ [0.006, 0.014]; $n=203$ & 2.05e-07 \\
\addlinespace
\bottomrule
\end{tabular}

\end{table}

\FloatBarrier
\section{Expanded Per-Dataset Explanations}

\begin{figure}[H]
\centering
\includegraphics[width=\textwidth]{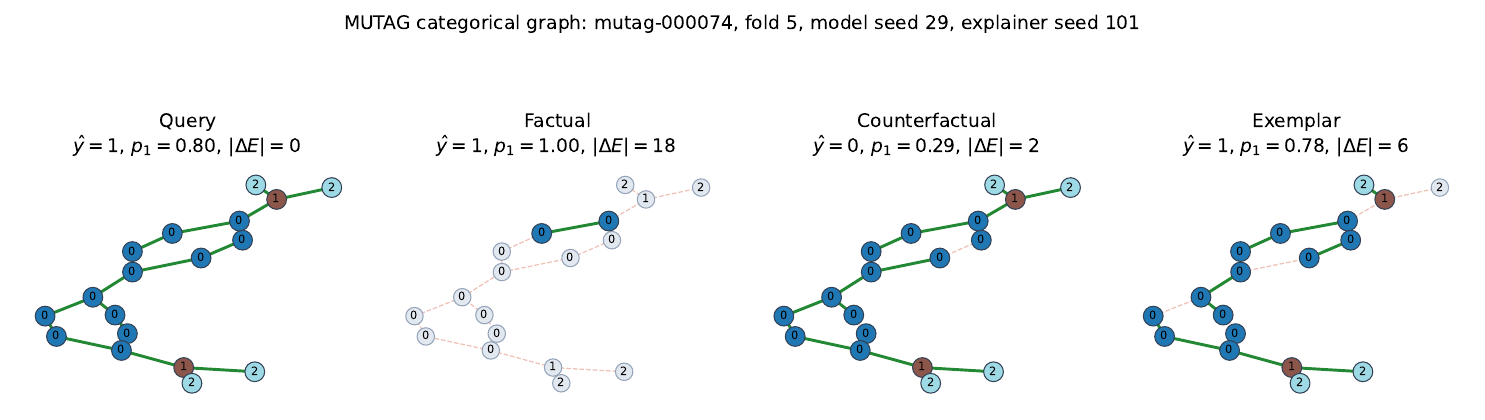}
\caption{Expanded MUTAG query, factual, counterfactual, and exemplar case.}
\label{fig:strip-mutag}
\end{figure}

\begin{figure}[H]
\centering
\includegraphics[width=\textwidth]{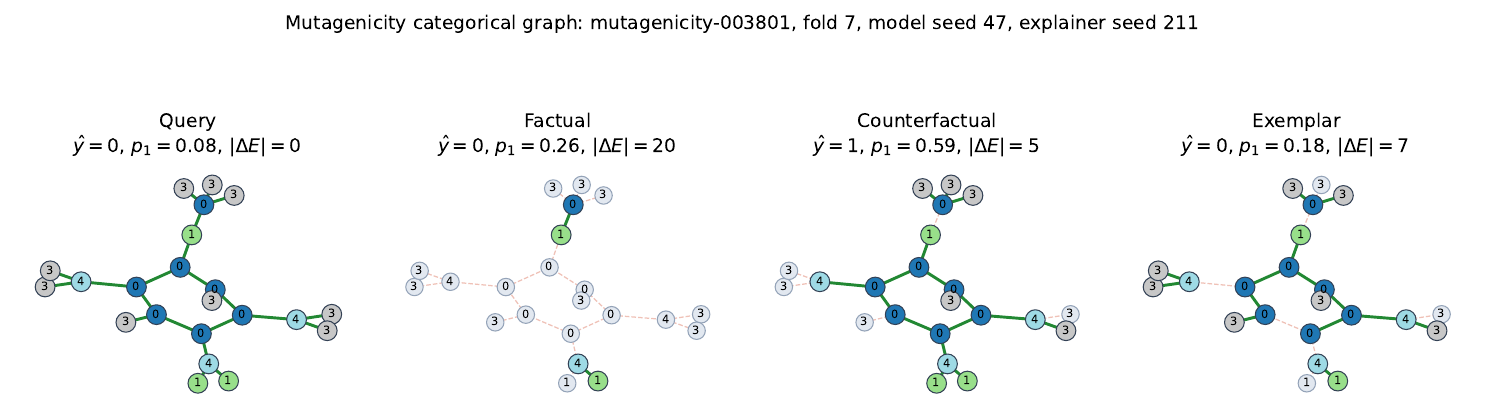}
\caption{Expanded Mutagenicity query, factual, counterfactual, and exemplar case.}
\label{fig:strip-mutagenicity}
\end{figure}

\begin{figure}[H]
\centering
\includegraphics[width=\textwidth]{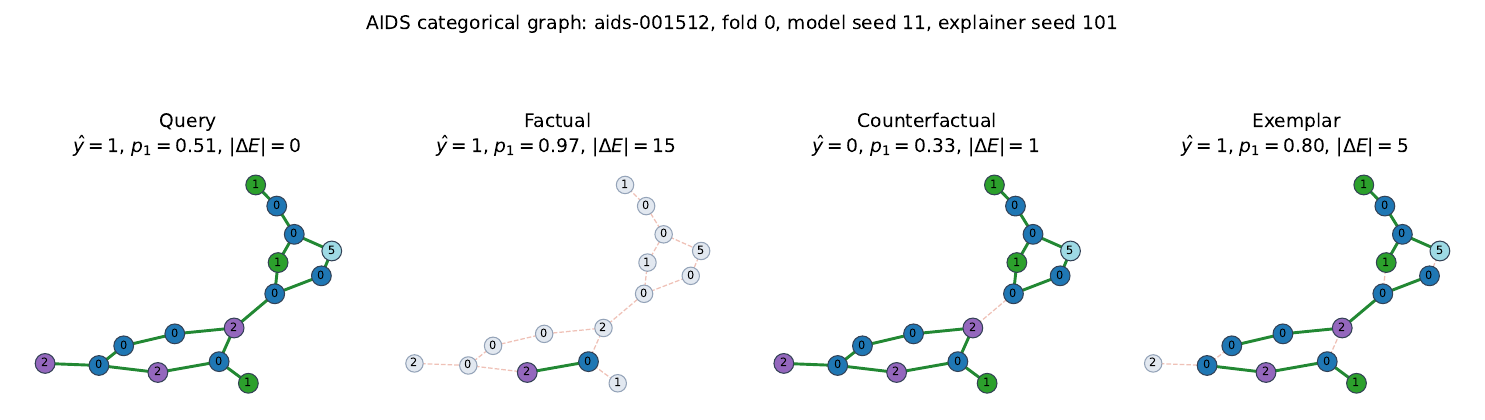}
\caption{Expanded AIDS query, factual, counterfactual, and exemplar case.}
\label{fig:strip-aids}
\end{figure}

\begin{figure}[H]
\centering
\includegraphics[width=\textwidth]{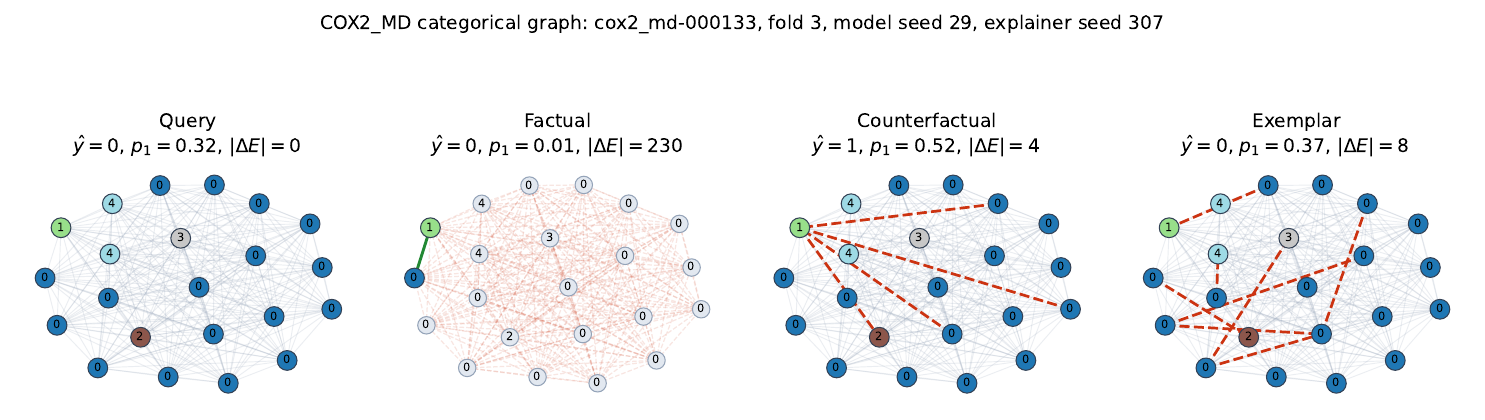}
\caption{Expanded COX2\_MD query, factual, counterfactual, and exemplar case.
The light complete-graph context and red dashed edits prevent the dense input
from obscuring the selected changes.}
\label{fig:strip-cox2}
\end{figure}

\begin{figure}[H]
\centering
\includegraphics[width=\textwidth]{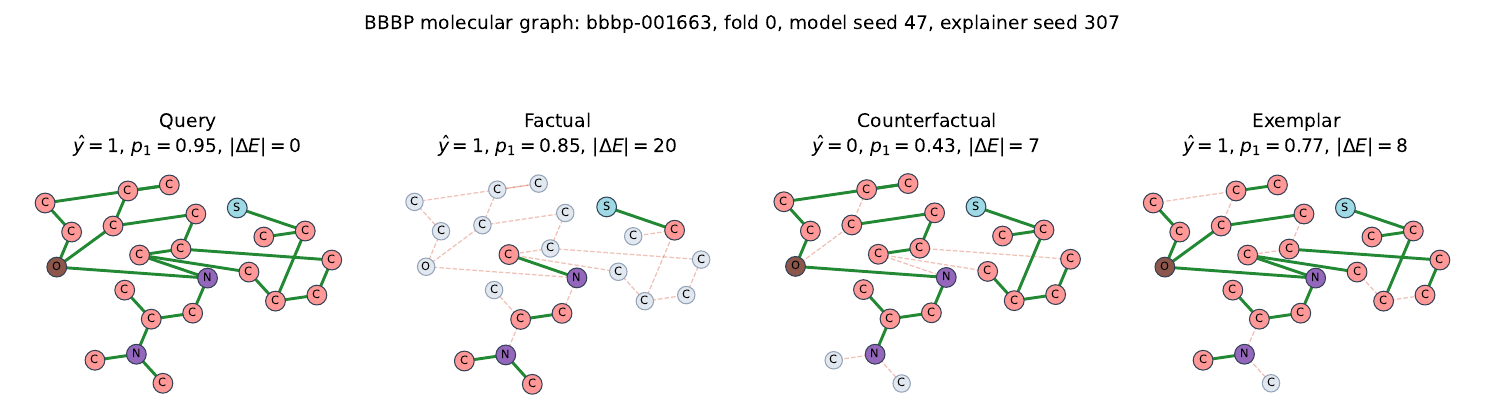}
\caption{Expanded BBBP query, factual, counterfactual, and exemplar edge-mask
case using RDKit coordinates.}
\label{fig:strip-bbbp}
\end{figure}

Figures~\ref{fig:strip-mutag}--\ref{fig:strip-bbbp} enlarge the five rows of
the main qualitative overview. Each strip uses one query graph and presents
the factual, counterfactual, and exemplar outputs in the same left-to-right
grammar as the primary BBBP fragment example. All evaluated query nodes
remain visible; grey nodes have no retained incident edge but their features
are still supplied to the predictor. Red dashed edges are removed and green
edges retained. COX2\_MD is a dense distance graph, so retained context is faint and
the comparatively few removed edges are red and dashed. TU node labels denote
dataset categories, not inferred chemical elements. The BBBP strip uses RDKit
coordinates for its recorded SMILES but remains an edge-mask explanation; it
is distinct from the separately reported sanitized fragment search. The exact
primary sanitized-fragment rendering is shown in
Fig.~\ref{fig:bbbp-verified-fragments}.

\FloatBarrier
\section{Reproducibility and Failure Records}

Every record is keyed by dataset, fold, architecture, predictor seed,
explainer seed, graph ID, method, perspective, budget, and protocol hash.
Atomic shards permit interrupted or parallel execution without silently
overwriting completed cells. A failed explanation remains a row with its
violated constraint and runtime. The claim-to-artifact map in the result root
links historical generated artifacts to their sources. The submission-specific
\texttt{manuscript\_claim\_to\_artifact\_map.csv} instead lists only referenced
artifacts, their source records, and SHA-256 hashes, excluding unused pilot
BBBP and D4 groups. The primary BBBP selection CSV and revised full-node
panel manifest retain figure provenance. A separate
\texttt{claim\_support\_audit.csv} records the scientific claim checks.

\begin{table}[H]
\centering
\caption{Failure audit for the primary GCN and proposed method. Failure rate
is the percentage of all attempts; the final column counts the most common
machine-readable reason.}
\scriptsize
\begin{tabular}{llrll}
\toprule
Dataset & View & Failure rate & Most common reason & Count \\
\midrule
MUTAG & Factual & 0.0\% & -- & -- \\
MUTAG & Counterfactual & 32.4\% & No Flip Within 8 Edits & 173 \\
MUTAG & Exemplar & 1.1\% & No Nontrivial Preserving Edit Within 3 Edits & 9 \\
\addlinespace
Mutagenicity & Factual & 0.0\% & -- & -- \\
Mutagenicity & Counterfactual & 36.3\% & No Flip Within 8 Edits & 10406 \\
Mutagenicity & Exemplar & 1.0\% & No Nontrivial Preserving Edit Within 8 Edits & 95 \\
\addlinespace
AIDS & Factual & 0.0\% & -- & -- \\
AIDS & Counterfactual & 95.2\% & No Flip Within 3 Edits & 5811 \\
AIDS & Exemplar & 0.0\% & No Nontrivial Preserving Edit Within 4 Edits & 3 \\
\addlinespace
COX2\_MD & Factual & 0.0\% & -- & -- \\
COX2\_MD & Counterfactual & 94.5\% & No Flip Within 8 Edits & 2578 \\
COX2\_MD & Exemplar & 0.0\% & -- & -- \\
\addlinespace
BBBP & Factual & 0.0\% & -- & -- \\
BBBP & Counterfactual & 45.4\% & No Flip Within 8 Edits & 664 \\
BBBP & Exemplar & 0.3\% & No Nontrivial Preserving Edit Within 3 Edits & 2 \\
\addlinespace
\bottomrule
\end{tabular}

\end{table}

\clearpage
\bibliographystyle{IEEEtran}
\bibliography{paper_references}
\end{document}